\documentclass[]{opendatalab}
\usepackage{xcolor}
\usepackage{pifont}
\usepackage[misc]{ifsym}
\usepackage{longtable}
\usepackage{listings}
\usepackage{wrapfig}
\usepackage[numbers]{natbib}
\usepackage{enumitem}
\usepackage{arydshln}
\usepackage[utf8]{inputenc}
\usepackage[T1]{fontenc}
\usepackage{xcolor,xstring}
\usepackage{fontawesome5} 
\definecolor{myred}{HTML}{A83B27}
\definecolor{mygreen}{HTML}{44985C}
\usepackage[table]{xcolor} 
\usepackage{booktabs}      
\usepackage{graphicx}      
\tcbuselibrary{listings, minted}

\usepackage{graphicx}         
\usepackage{booktabs}         
\usepackage{tabularray}
\UseTblrLibrary{booktabs} 
\usepackage[table]{xcolor}    
\usepackage{xstring}          

\makeatletter
\newcommand\andauthor{%
  \g@addto@macro\authorlist{\par\vspace{1mm}}
}

\makeatother

\definecolor{groupcolor}{gray}{0.92}
\definecolor{mygreen}{RGB}{40, 160, 40}
\definecolor{myred}{RGB}{200, 30, 30}

\definecolor{codegreen}{rgb}{0,0.6,0}
\definecolor{codegray}{rgb}{0.5,0.5,0.5}
\definecolor{codepurple}{rgb}{0.58,0,0.82}
\definecolor{backcolour}{rgb}{0.95,0.95,0.92}
\definecolor{promptcolor}{HTML}{D1D0F2}
\definecolor{promptcolorheader}{HTML}{bdbcec}
\newcommand{\promptbox}[2]{
\begin{tcolorbox}[
top=0.3em,bottom=0.3em,left=0.5em,right=0.5em,
toptitle=0.3em,bottomtitle=0.2em,boxsep=0pt,
colframe=promptcolorheader,colback=promptcolor!50,boxrule=0.5pt,
]
\footnotesize
\end{tcolorbox}
}

\newcommand{\deltaval}[1]{%
  \text{\scriptsize
    \IfBeginWith{#1}{-}{\textcolor{myred}{(#1)}}{\textcolor{mygreen}{(#1)}}%
  }%
}
\newcommand{\valdelta}[2]{%
  #1\makebox[7mm][r]{\deltaval{#2}}%
}
\usepackage{adjustbox}
\usepackage{booktabs} 
\usepackage[T1]{fontenc}
\tcbset{
  takeawaysbox/.style={
    title=Takeaway, 
    colback=teal!5!white,
    colframe=teal!65!black,
    fonttitle=\bfseries\sffamily\scshape\small,
    coltitle=white,
    colbacktitle=teal!50!black,
    enhanced,
    attach boxed title to top left={xshift=2mm, yshifttext=-1mm, yshift=-2mm},
    boxed title style={
        outer arc=2mm,
        arc=1.5mm, 
        colframe=teal!65!black,
        colback=teal!50!black,
    },
    width=\linewidth,
    arc=3mm,
    boxrule=0.8pt, 
  }
}

\lstdefinestyle{mystyle}{
    backgroundcolor=\color{backcolour},   
    commentstyle=\color{codegreen},
    keywordstyle=\color{magenta},
    numberstyle=\tiny\color{codegray},
    stringstyle=\color{codepurple},
    basicstyle=\ttfamily\footnotesize,
    breakatwhitespace=false,         
    breaklines=true,                 
    captionpos=b,                    
    keepspaces=true,                 
    numbers=left,                    
    numbersep=5pt,                  
    showspaces=false,                
    showstringspaces=false,
    showtabs=false,                  
    tabsize=2
}

\definecolor{myLavender}{HTML}{e2d5ba}
\definecolor{myDarkBlue}{HTML}{14bc94}
\definecolor{green1}{HTML}{f3eedd}
\definecolor{purple1}{HTML}{303030}
\definecolor{green2}{HTML}{BFF6BA}
\definecolor{blue2}{HTML}{508AB2}
\definecolor{red1}{HTML}{ff6600}
\definecolor{blue1}{HTML}{0085c7}

\newtcolorbox{findbox}[2][]{
    enhanced,
    colback=green2!18,
    colframe=myDarkBlue!50,
    arc=3pt,
    title=#2,
    fonttitle=\bfseries\large,
    coltitle=blue,
    colbacktitle=myDarkBlue!60,
    attach boxed title to top center={yshift=-2mm},
    boxed title style={
        colframe=myDarkBlue,
        arc=3pt,
    },
    coltext=black!85,
    fontupper=\linespread{1.2}\selectfont,
    #1 
}

\newtcbtheorem[number within=section]{exmp}{Example}%
{beforeafter skip balanced=.01in,colback=green1!30,colframe=purple1!70,fonttitle=\small \bfseries, left=.03in, right=.03in,bottom=.02in, top=.02in}{exmp}

\title{SDAR: A Synergistic Diffusion-AutoRegression Paradigm for Scalable Sequence Generation}

\author[1,2*]{Shuang Cheng}
\author[3*]{Yihan Bian}
\author[1,4*]{Dawei Liu}
\author[4]{Linfeng Zhang}
\author[1]{Qian Yao}
\author[1]{Zhongbo Tian}
\author[1]{Wenhai Wang}
\author[1]{Qipeng Guo}
\author[1]{Kai Chen}
\author[1\ddagger\textrm{\Letter}]{Biqing Qi}
\author[1,5\textrm{\Letter}]{Bowen Zhou}

\affiliation[1]{Shanghai AI Laboratory}
\affiliation[2]{Zhejiang University}
\affiliation[3]{University of Maryland, College Park}
\affiliation[4]{Shanghai Jiao Tong University}
\affiliation[5]{Tsinghua University}

\abstract{
We propose \textsc{SDAR}, a \textbf{S}ynergistic \textbf{D}iffusion–\textbf{A}uto\textbf{R}egression paradigm that establishes a new language modeling framework combining the training efficiency of autoregression with the parallel inference capability of diffusion. Instead of costly end-to-end diffusion training, SDAR performs a lightweight paradigm conversion that transforms a well-trained autoregressive (AR) model into a blockwise diffusion model through brief, data-efficient adaptation. During inference, SDAR models generate sequences autoregressively across blocks for global coherence while decoding all tokens within each block in parallel via a discrete diffusion process. Through extensive controlled experiments, we demonstrate that AR models remain substantially more compute-efficient than masked diffusion models, providing a strong foundation for adaptation. Building on this insight, SDAR achieves efficient AR-to-diffusion conversion with minimal cost, preserving AR-level performance while enabling parallel generation. Scaling studies across both dense and Mixture-of-Experts architectures further confirm that SDAR scales without compromise—larger models exhibit increasing robustness to block size and decoding thresholds, yielding greater parallel speedups without loss of accuracy. Beyond efficiency, SDAR also exhibits enhanced reasoning and domain adaptability. Our 30B MoE model surpasses its AR counterpart on challenging scientific reasoning benchmarks such as GPQA and ChemBench, benefiting from local bidirectional context and reduced causal constraints. When combined with test-time scaling strategies such as majority voting and pass@k, SDAR achieves substantial additional gains, indicating strong potential for reinforcement learning optimization. Together, these results establish SDAR as a new and practical language modeling paradigm that unifies the complementary strengths of autoregression and diffusion, enabling scalable, high-throughput inference while preserving the accuracy and reasoning competence of state-of-the-art AR models.
}

\metadata[* Equal contribution\quad $\textrm{\Letter}$ Corresponding author \quad $\ddagger$ Project leader]{}
\metadata[Github:]{\url{https://github.com/JetAstra/SDAR}}
\metadata[Huggingface:]{\url{https://huggingface.co/JetLM}}

\begin{document}

\maketitle

\section{Introduction}
Large Language Models (LLMs) are primarily built upon the autoregressive (AR) paradigm, modeling sequences from left to right via next-token prediction~\citep{achiam2023gpt,grattafiori2024llama,guo2025deepseek,bengio2003neural,radford2018improving,radford2019language}. This approach imposes a strict, token-level causal inductive bias, which aligns naturally with the sequential structure of natural language. 
However, the very success of this causal inductive bias creates a fundamental tension. First, the strictly sequential nature of AR decoding—generating token by token—imposes a severe bottleneck on inference speed, precluding parallelization, thereby increasing latency and serving costs at scale.
Second, this token-level causal dependency creates a misalignment with tasks that demand non-local or holistic reasoning, particularly in scientific problems where bidirectional attention facilitates the identification of chemical functional groups, DNA sequences, or molecular formulas. For these tasks, the rigid left-to-right generation process can be detrimental, necessitating substantially greater compute and data to mitigate the limitations of sequential decoding.

As an alternative, masked diffusion language models (MDLMs) have recently become the most widely adopted paradigm among diffusion-based LMs, proposing a fundamentally different philosophy. By treating the entire sequence as a holistic entity to be generated jointly, these models circumvent the strict causal constraints of autoregressive methods, enabling a more flexible generation process with advantages such as arbitrary generation order and parallel decoding~\citep{li2025survey, yu2025discrete, gulrajani2023likelihood, sahoo2024simple}.

However, MDLMs face two fundamental challenges that limit their training and inference efficiency. First, they typically optimize the Evidence Lower Bound (ELBO)~\citep{lou2024discrete, gongscaling, nie2025large}, rather than the standard negative log-likelihood. Since NELBO is only a loose upper bound of NLL, this objective is inherently less efficient~\citep{nie2024scaling, arriola2025block, feng2025theoretical}, and empirical studies—both ours and others’—consistently validate a substantial performance gap compared to autoregressive models trained directly on NLL. Second, current open-source implementations~\citep{nie2025large, ye2025dream} incur severe inference costs. The absence of KV caching, coupled with the need to avoid quality degradation, results in an $O(N^3)$ computational complexity per sequence. While some works explore approximate KV caching for diffusion models, it remains unclear whether such methods can resolve the bottleneck without harming other capabilities.

To reconcile the trade-offs between autoregressive and diffusion-based models, a class of hybrid models has been explored to unify the diffusion and autoregressive approaches~\citep{han2022ssd, arriola2025block, fathi2025unifying}. The core mechanism involves a block-wise decomposition of the sequence. At a local level, diffusion models handle intra-block generation in parallel, thereby relaxing the strict causal constraints. At a global level, an autoregressive framework models the dependencies between these blocks. This hierarchical strategy advantageously retains the macroscopic causal structure (or "Global AR-ness" \cite{gong2025diffucoder}) of language, which not only tights the learning objective but also preserves practical functionalities such as variable-length generation and KV-caching. Simultaneously, it alleviates the suboptimal inductive bias of pure AR models on locally complex dependencies, unlocking substantial gains in modeling accuracy and decoding speed.

However, the practical viability of these hybrid architectures is severely hampered by fundamental challenges in training efficiency. The diffusion component, often optimized with slow-converging objectives like the ELBO, incurs a prohibitive computational cost. Empirical studies show that MDLM require substantially more FLOPs to match the performence of their autoregressive counterparts, and often demand prolonged training. Compounding this issue, the hybrid block-wise AR-diffusion strategy itself introduces a significant training overhead, with a composite objective that can nearly double the computational expenditure per instance compared to a pure AR or diffusion strategy~\citep{arriola2025block}.

Building on these insights, we design {SDAR} to reconcile the efficiency of autoregressive training with the parallelism of diffusion-based inference. The key principle is \emph{decoupling} the two phases: we leverage full-scale AR pretraining to ensure stability and efficiency, and then introduce a lightweight adaptation stage that equips the model with block-wise diffusion decoding. This design preserves the practical advantages of AR—such as KV caching, variable-length generation, and strong optimization behavior—while unlocking diffusion’s unique benefit of parallel intra-block generation.

To evaluate this paradigm, we conduct two parts of experiments. First, we perform a controlled comparison between AR and masked diffusion language models (MDLMs) to quantify training efficiency under identical computational settings. Based on that, we examine the feasibility of adapting SDAR from both AR and MDLM bases, revealing that AR backbones consistently provide a stronger foundation. Second, we scale SDAR across dense and MoE architectures, systematically analyzing the trade-offs between model size, block size, training cost, and inference speed. Together, these studies provide the first comprehensive assessment of hybrid AR–diffusion language modeling and establish SDAR as a practical recipe for the community.

In summary, our work makes the following contributions:
\begin{enumerate}
	\item \textbf{Fair comparison of AR and MDLM training efficiency.} We conduct controlled experiments comparing autoregressive (AR) models with masked diffusion language models (MDLMs), using the same training data and recommended hyperparameters. This provides the first real-world, meaningful evidence that AR models deliver substantially higher training efficiency than MDLMs under identical computational settings.
	\item \textbf{Adaptation of SDAR from AR and MDLM bases.} We systematically evaluate the feasibility of adapting SDAR from both AR and MDLM backbones. Using a 100B-token subset of the 1T training corpus, we perform thorough downstream task benchmarks, demonstrating that AR-based adaptation consistently yields superior performance and is therefore the preferred choice for serving as the base model.
	\item \textbf{Scaling experiments and adaptation recipe.} We extend our study to larger models by adapting state-of-the-art open-source AR models (both dense and MoE variants) using an arbitrary 50B-token open-source dataset. We provide a practical recipe showing that any AR decoder-only model, regardless of its architecture, can be efficiently adapted into SDAR—even without access to its original pretraining data. We argue that this adaptation requires significantly less data than adaptation method such as Dream, making it broadly accessible to the community.
	\item \textbf{Comprehensive analysis of scaling laws.} We offer detailed empirical analysis of the relationship between model size, block size, downstream task performence and inference speed. Our study spans models from 1.7B to 30B parameters, covering both dense and MoE architectures, and block sizes ranging from 4 to 64. We also show that with minimal additional supervised fine-tuning, block size can be increased from a base SDAR model at negligible cost.
	\item \textbf{Open-sourced models and inference engine.} We release all trained SDAR models across different sizes(1.7B, 4B, 8B, 30B) and block configurations(from 4 to 64), along with both research-oriented and production-ready inference engines. For research and educational purposes, we provide JetEngine, a lightweight implementation prioritizing code clarity. For industrial applications, we offer a high-performance implementation on the lmdeploy framework with block-oriented optimizations for efficient SDAR inference. Our release also includes SDAR-30B-A3B-Sci, the most powerful diffusion-based reasoning model to date, and the first diffusion model capable of long chain-of-thought generation (see Appendix ~\ref{appendix:exp}), achieving state-of-the-art performance in complex reasoning tasks.
\end{enumerate}


\section{Preliminary: Language Modeling Paradigm}
\label{sec:preliminary}

We begin by establishing our notation and reviewing the foundational paradigms in language modeling that are pertinent to our work.
We consider a vocabulary $\mathcal{V}$ of size $V$. A token at position $\ell$ is represented by its index $x^\ell \in \mathcal{V}$, and a sequence of length $L$ is a tuple $x = (x^1, \dots, x^L)$.
For computational purposes, we represent each token $x^\ell$ by its one-hot vector $\mathbf{x}^\ell \in \{0, 1\}^V$. The entire sequence is thus represented by $\mathbf{x} = (\mathbf{x}^1, \dots, \mathbf{x}^L)$.
Finally, we denote the probability simplex over the vocabulary as $\Delta^V$, and use $\text{Cat}(\cdot; p)$ for a categorical distribution with probabilities $p \in \Delta^V$.

\subsection{Autoregressive Models}
\label{ssec:ar}

Autoregressive (AR) models model the probability of a sequence $x$ by decomposing the joint distribution into a product of conditional probabilities:
\begin{equation}
\label{eq:ar}
\log p_{\theta}(x) = \sum_{\ell=1}^{L} \log p_{\theta}(x^{\ell} | x^{<\ell})
\end{equation}
where $x^{<\ell}$ is the sequence of preceding tokens. The models are trained with maximum likelihood estimation on the next-token prediction task. The sequential nature of this process, however, imposes that generation is iterative; generating a sequence of length $L$ requires $L$ sequential forward passes, which is a major bottleneck for low-latency applications.

\subsection{Discrete Masked Denoising Diffusion Models}
\label{ssec:dmddm}

Discrete masked diffusion models are non-autoregressive generative models that learn to reverse a corruption process \citep{austin2021structured}. This forward process is typically applied independently to each token.
The conditional distribution $q(x_t^\ell|x_0^\ell)$ is a categorical distribution, parameterized by the linear interpolation of the original token's one-hot vector $\mathbf{x}_0^\ell$ and the one-hot vector $\mathbf{m}$ for the special [MASK] token.
The interpolation coefficient $\alpha_t$ is governed by a predefined noise schedule, such as a linear one $\alpha_t = 1-t$~\citep{nie2025large,ye2025dream}:
\begin{equation}
\label{eq:forward_process_corrected}
q(x_t^\ell|x_0^\ell) = \text{Cat}(x_t^\ell; \alpha_t\mathbf{x}_0^\ell + (1-\alpha_t)\mathbf{m})
\end{equation}

The generative model $p_{\theta}$ is trained to reverse this process by minimizing the negative evidence lower bound objective (NELBO). For this absorbing-state formulation, the NELBO simplifies to a reweighted cross-entropy objective:

\begin{equation}
\label{eq:diffusion_loss}
\mathcal{L}(\theta) = \mathbb{E}_{x_0 \sim p_{\text{data}}, {x}_t \sim q({x}_t|{x}_0), t \sim U(0,1)} \left[-\frac{1}{t} \sum_{\ell=1}^{L} \mathbf{1}[{x}_t^\ell = \text{[MASK]}] \log p_{\theta}({x}_0^\ell | {x}_t) \right]
\end{equation}
where the indicator function $\mathbf{1}[{x}_t^\ell = \text{[MASK]}]$ ensures the loss is computed only on tokens replaced by the absorbing state [MASK]. The term $1/t$ is a time-dependent weighting factor derived from the diffusion dynamics under the linear noise schedule $\alpha_t = 1-t$.

\subsection{Blockwise Diffusion Language Models}
\label{ssec:bdlm}

To enable variable-length generation and the KV cache mechanism for diffusion models, several works have proposed blockwise diffusion frameworks (also known as semi-autoregressive diffusion frameworks) \citep{fathi2025unifying, arriola2025block, han2022ssd}.
This hybrid approach partitions a sequence $x$ into $B$ contiguous, non-overlapping blocks, $\{x^1, \dots, x^B\}$, with each block containing $L'$ tokens. The modeling is autoregressive at the block level while being non-autoregressive within each block:
\begin{equation}\label{eq:blockwise}
\log p_{\theta}(x) = \sum_{b=1}^{B} \log p_{\theta}(x^{b} | x^{<b})
\end{equation}

Each conditional distribution $p_{\theta}(x^{b} | x^{<b})$ is modeled by a self-contained diffusion process. When applying the \textit{Discrete Masked Denoising Diffusion} strategy (Section~\ref{ssec:dmddm}), the training objective becomes the minimization of the expected conditional NELBO over blocks:
\begin{equation}
\label{eq:blockwise_loss}
\mathcal{L}_{\text{blockwise}}(\theta) = \mathbb{E}_{x \sim p_{\text{data}}, b \sim U[1, B], t} \left[-\frac{1}{t} \sum_{\ell=1}^{L'} \mathbf{1}[{x}_t^{b,\ell} = \text{[MASK]}] \log p_{\theta}({x}_0^{b,\ell} | {x}_t^b, x^{<b}) \right]
\end{equation}
where $L'$ is the block size, ${x}_t^b$ is the noised sequence for block $b$, and ${x}_0^{b,\ell}$ denotes the $\ell$-th original token in that block. The model is thus trained to reconstruct the original block $x^b$ from its corrupted version ${x}_t^b$, conditioned on the preceding clean blocks $x^{<b}$.
\section{Method}

The core of \textsc{SDAR} lies in a decoupled paradigm that combines the training efficiency of autoregressive (AR) modeling with the parallel inference capability of diffusion. Instead of performing costly end-to-end block-diffusion training, \textsc{SDAR} leverages a well-trained AR model as a foundation and introduces a highly efficient adaptation procedure that transforms it into a block-wise diffusion model. This process enables parallel generation while preserving the AR model’s strong language understanding and reasoning capabilities. After adaptation, a standard instruction-based supervised fine-tuning (SFT) stage is applied to align the model with human preferences, thereby endowing it with conversational and reasoning abilities at minimal additional cost.

\subsection{The \textsc{SDAR} Training Paradigm}

Starting from a sufficiently capable AR base model $\theta_{\text{AR}}$ via conventional NTP pretraining, we subsequently continue training the model to convert the language modeling paradigm from AR to block-wise diffusion. This strategy avoids the prohibitive computational cost of training a block-wise diffusion model from scratch, which is markedly less efficient than AR training.

\begin{figure}[htbp]
    \centering
    \includegraphics[width=0.98\textwidth]{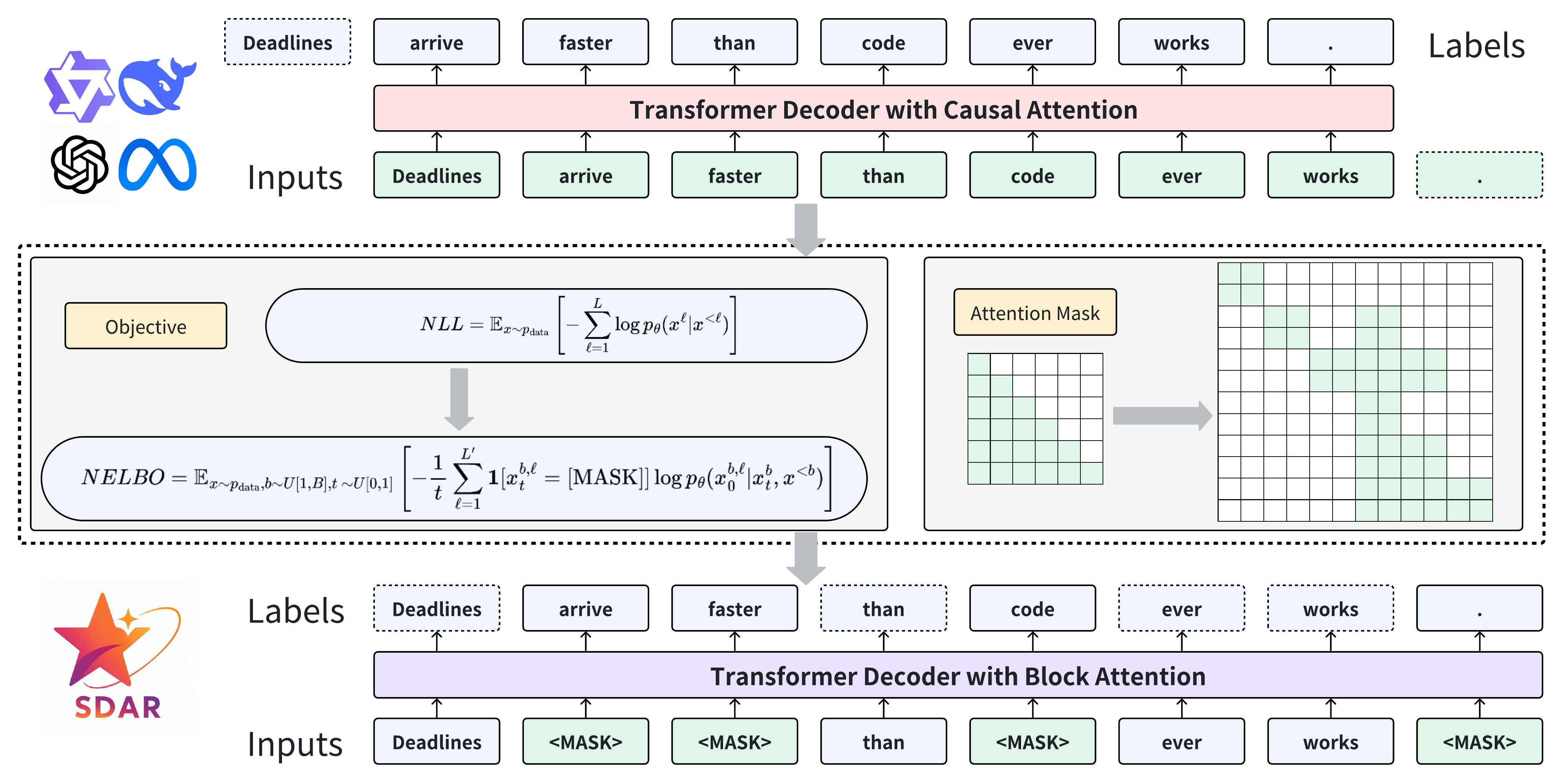}
    \caption{The training paradigm of converting AR to SDAR. First the traditional NTP training is adopted to obtain the AR base model. We then continue training the AR model by modifying the attention mask and replacing the training objective from NLL to NELBO, converting it into a block-wise diffusion model (SDAR), without logits shift or attention mask annealing.}
    \label{fig:training}
\end{figure} 

We first perturb the sequence by randomly masking tokens following the MDLM procedure. The entire sequence is perturbed during the pretraining stage, whereas during SFT, the prompt remains clean and the response is perturbed. We partition a sequence $x$ into $K$ non-overlapping blocks of size $B$, such that $x = (b_1, b_2, \dots, b_K)$, where each block $b_k = (x_{(k-1)B+1}, \dots, x_{kB})$. The modeling objective now becomes a hybrid of inter-block autoregression and intra-block parallel diffusion, as defined in the block-wise diffusion preliminary. The probability of a sequence is factorized as:
\begin{equation}
P(x | \theta) = \prod_{k=1}^{K} P(b_k | b_{<k}; \theta).
\end{equation}
Each conditional term $P(b_k | b_{<k}; \theta)$ is modeled using a discrete diffusion process. Specifically, during this training stage, the model (initialized with $\theta_{\text{AR}}$) learns to denoise a corrupted block $b_k^t$ at diffusion step $t$, conditioned on the preceding clean blocks $b_{<k}$. The clean blocks adopt block-wise causal attention, while corrupted blocks attend to themselves and all preceding blocks. As illustrated in Figure~\ref{fig:training}, this can be achieved by concatenating the perturbed and clean sequences into a single input for the forward pass, with the attention mask modified accordingly~\citep{arriola2025block}. The objective is to optimize the conditional block-wise ELBO instead of log-likelihood, as introduced in the preliminaries:
\begin{equation}
\mathcal{L}_{\text{blockwise}}(\theta) = \mathbb{E}_{t, q(b_k^t|b_k^0)} \left[ \log P(b_k^0 | b_k^t, b_{<k}; \theta) \right].
\end{equation}
The conversion process does not involve logits shift or attention mask annealing. Crucially, this conversion is performed on a significantly smaller dataset (e.g., $\text{30 B}\sim\text{50 B}$ tokens, compared to trillions for pretraining). The pretrained AR model provides a powerful initialization, allowing for rapid convergence on the new objective. This step effectively "unlocks" the model's latent ability to perform holistic, non-causal reasoning within local blocks, preparing it for parallel inference.

\subsection{Hierarchical Inference and Decoding Strategies}
\begin{figure}[htbp]
    \centering
    \includegraphics[width=0.95\textwidth]{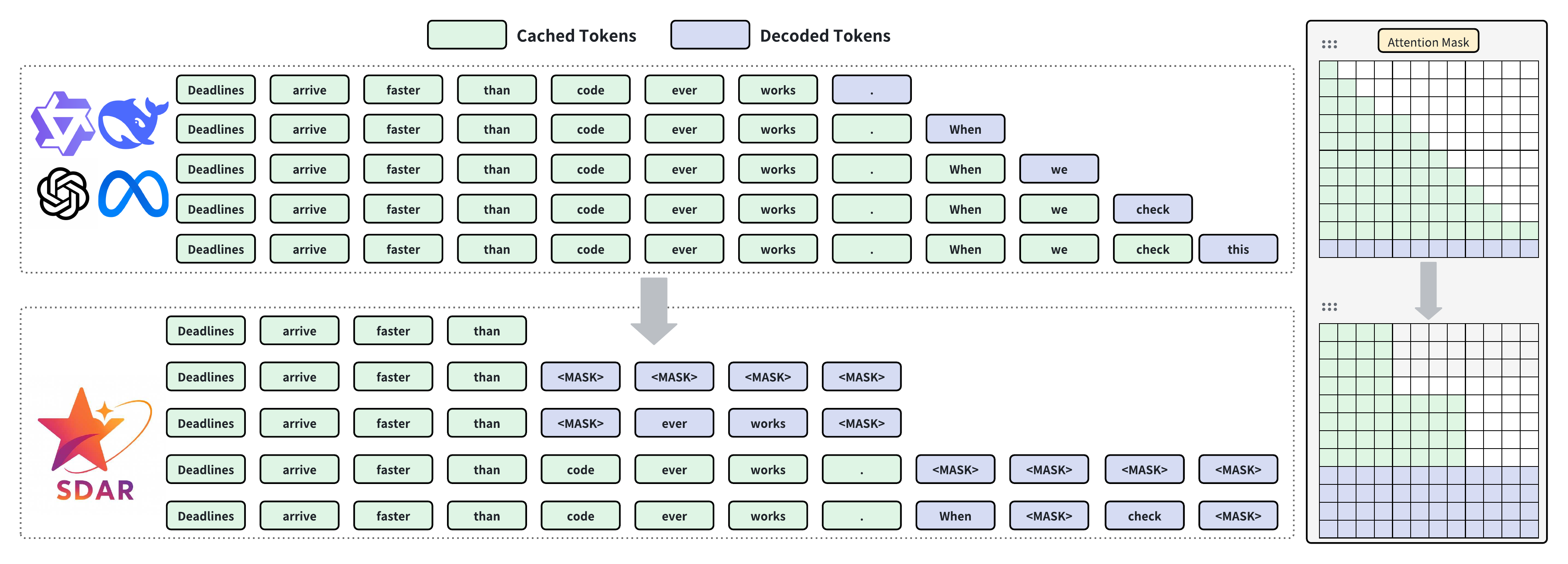}
    \caption{Contrasting inference paradigms between AR and SDAR. SDAR adopts a block-wise causal attention mechanism to enable inter-block causal autoregression and intra-block parallel diffusion. The KV cache from previously generated blocks is reused, while the block currently being decoded does not store KV cache.}
    \label{fig:inference}
\end{figure} 

The inference process of \textsc{SDAR} is hierarchical: it generates text autoregressively at the block level while employing a parallel diffusion-based mechanism for decoding within each block. This hybrid approach marries the long-range coherence of AR models with the speed of parallel generation.

Given a prompt, sequence generation proceeds by sequentially producing blocks $b_1, b_2, \dots, b_K$. The generation of each block $b_k$ is conditioned on the entire preceding context, which consists of the initial prompt followed by any previously generated blocks ($b_1, \dots, b_{k-1}$).

To generate a single block $b_k$, we initiate a parallel decoding process starting from a sequence of $B$ `[MASK]` tokens, which we denote as the fully-noised state $b_k^T$. The model then iteratively refines this block over $T$ denoising steps. At each step $t$ (from $T$ down to 1), the model takes the current state of the block $b_k^t$ and the preceding context to predict the logits for the less noisy block $b_k^{t-1}$:
\begin{equation}
\text{logits}(b_k^{t-1}) \sim P(\cdot | b_k^t, b_{k-1},\text{prompt}; \theta).
\end{equation}
From the resulting logits, a subset of the masked positions is selected for decoding. For each selected position, a token is then generated from its corresponding probability distribution using a standard decoding strategy (e.g., greedy search, nucleus sampling). The remaining positions are kept as `[MASK]` tokens, forming the input $b_k^{t-1}$ for the next iteration. This denoising loop continues until all positions in the block are filled, at which point the model proceeds to generate the next block, $b_{k+1}$.
To determine which positions to decode at each step, we propose two primary remasking strategies:

\paragraph{Static Low Confidence Remasking.} This strategy decodes a fixed number of tokens at each step. For a block of size $B$ and a total of $T$ denoising steps, we select the $\lceil B/T \rceil$ masked positions with the highest prediction confidence (i.e., the highest probability assigned to the most likely token) to be decoded in the current step. This approach is simple and predictable, ensuring the generation of a block completes in a constant number of forward passes.

\paragraph{Dynamic Low Confidence Remasking.} This adaptive strategy accelerates decoding by finalizing tokens based on model confidence. At each step, we select any masked position where the prediction confidence exceeds a predefined threshold $\tau$. To guarantee progress, a minimum number of positions (e.g., one, or $\lceil B/T \rceil$) are always selected based on highest confidence if the threshold is not met by a sufficient number of candidates. This allows the model to fill in "easy" parts of a block in fewer steps, reducing the overall number of required forward passes for generation.

\section{From AR and Diffusion to SDAR}\label{part1}

\begin{tcolorbox}[takeawaysbox]
(1) \textbf{AR models are more compute-efficient} than masked diffusion objectives, making them the stronger foundational choice under equal budgets.

(2) \textbf{SDAR enables efficient AR-to-Block Diffusion conversion}, preserving AR performance while adding parallel decoding.

(3) \textbf{AR backbones remain superior post-conversion}, with AR-derived SDAR models consistently outperforming MDLM-based ones.
\end{tcolorbox}

Before introducing SDAR, we begin by revisiting the fundamental comparison between Autoregressive (AR) and Masked Diffusion Language Models (MDLMs). Under identical architectures, datasets, and training budgets, AR models consistently achieve lower training loss and better downstream performance than MDLMs. This efficiency arises because AR training directly optimizes the exact negative log-likelihood (NLL) through a cross-entropy objective, where every token in the sequence contributes to parameter updates, maximizing the utility of each sample. In contrast, MDLMs approximate this objective via a loose upper bound—the negative evidence lower bound (NELBO)—and rely on random masking, where only the masked subset of tokens contributes to the loss while the remaining unmasked tokens serve merely as context. Consequently, a substantial portion of the computation does not directly improve the model, resulting in slower convergence and higher FLOPs to reach comparable validation perplexity and downstream task performence.

This efficiency gap makes AR the natural foundation for practical large-scale training. However, AR’s left-to-right causality restricts parallel decoding, while MDLMs offer flexibility in generation order but lack KV caching and variable-length decoding, limiting real-world usability. Blockwise Diffusion (BD) provides a potential middle ground by enabling intra-block parallelism with block-level KV caching, but pretraining BD from scratch is prohibitively expensive, effectively doubling context length and compute requirements.

To resolve this tension, we propose {SDAR}, a lightweight conversion method that adapts an existing AR or MDLM backbone into BD through short continued training. SDAR retains AR’s superior training efficiency while inheriting BD’s parallel decoding capability, thereby combining the best of both paradigms without incurring the cost of full BD pretraining.

Our study therefore asks two controlled questions:
\begin{enumerate}
    \item Can an AR model be converted into a strong BD model at low cost, gaining parallelism without sacrificing quality?
    \item Between AR and MDLM backbones, which provides the stronger initialization for BD adaptation?
\end{enumerate}

\subsection{Experimental Setup}

\paragraph{Training Setup}
We pretrain two 2B-parameter models from scratch on an identical 1T-token corpus (general web, STEM, and code), using the same architecture, tokenizer, optimizer, and context length (4096 with packed sequences).
The AR model uses variable-length causal attention; the MDLM uses full bidirectional attention, which is standard and effective for diffusion-style training~\citep{nie2025large}.

After pretraining, we perform a 100B-token annealing training phase to obtain four Base models:
\begin{enumerate}
\item \textbf{AR-2B-Base:} The pre-trained AR model is continually trained with its original autoregressive objective.
\item \textbf{SDAR-2B-Base:} The pre-trained AR model is converted to the Block Diffusion paradigm, with a block size of 16. This is our proposed SDAR model.
\item \textbf{MDLM-2B-Base:} The pre-trained MDLM is continually trained with its original masked diffusion objective.
\item \textbf{MDLM-BD-2B-Base:} The pre-trained MDLM is converted to the Block Diffusion paradigm, with a block size of 16.
\end{enumerate}
Each Base model then undergoes supervised fine-tuning (SFT) on the same 4B-token instruction-following dataset using the same strategy as in its annealing phase, producing four chat variants: AR-2B-Chat, AR-BD-2B-Chat (SDAR-2B-Chat), MDLM-2B-Chat, and MDLM-BD-2B-Chat.

\paragraph{Evaluation Setup}
We evaluate AR-2B-Chat, AR-BD-2B-Chat, MDLM-2B-Chat, and MDLM-BD-2B-Chat on downstream tasks that probe reasoning, mathematics, and code:
\begin{itemize}
    \item \textbf{Reasoning \& Knowledge}: BBH~\citep{suzgun2022challenging} (3-shot), MMLU~\citep{hendryckstest2021} (5-shot), GPQA-Diamond~\citep{rein2024gpqa} (0-shot).
    \item \textbf{Mathematics}: MATH~\citep{lightman2023lets} (4-shot, CoT), GSM8K~\citep{cobbe2021gsm8k} (4-shot, CoT).
    \item \textbf{Coding Tasks}: HumanEval~\citep{chen2021codex} (0-shot), MBPP~\citep{austin2021program} (3-shot).
\end{itemize}

Our evaluation employs distinct decoding strategies matched to each model architecture. The AR-2B-Chat model uses standard greedy decoding. For our Block Diffusion variants, SDAR-2B-Chat and MDLM-BD-2B-Chat, we utilize a lowest-confidence static decoding method with a block length of 16 and 16 denoising steps. The MDLM-2B-Chat follows the LLaDA~\citep{nie2025large} protocol, using a lowest-confidence semi-autoregressive remasking approach with hyper-parameters set according to the original publication.

\subsection{Summary of Evaluation Results}
\begin{table}[h]
\centering
\caption{Benchmark Performance of 2B Chat Models. Our proposed SDAR model (\textbf{AR-BD}) successfully preserves the high performance of the original AR model, validating our low-cost conversion approach. It also consistently outperforms the MDLM-based counterparts on key reasoning benchmarks. All models are evaluated as chat variants.}
\label{tab:main_results_part1}
\begin{tabular}{lccccccc}
\toprule
\textbf{Model} & \textbf{BBH} & \textbf{MMLU} & \textbf{MATH} & \textbf{GSM8K} & \textbf{HumanEval} & \textbf{MBPP} & \textbf{GPQA} \\
\midrule
AR-2B-Chat & 35.7 & 48.7 & 29.9 & 61.8 & 42.1 & 44.4 & 26.3 \\
MDLM-2B-Chat & 32.2 & 47.0 & 12.6 & 57.9 & 21.3 & 27.2 & 26.3 \\
AR-BD-2B-Chat-b16 & 35.9 & 50.9 & 26.8 & 59.4 & 40.0 & 42.9 & 28.2 \\
MDLM-BD-2B-Chat-b16 & 32.9 & 47.5 & 23.3 & 64.7 & 39.0 & 33.5 & 29.8 \\
\bottomrule
\end{tabular}
\end{table}

The evaluation results in Table~\ref{tab:main_results_part1} reveal three key findings. 

\textbf{First}, there is a clear performance gap between the autoregressive and masked diffusion baselines. AR-2B-Chat substantially outperforms MDLM-2B-Chat across nearly all benchmarks, particularly in mathematics (MATH: 29.9 vs.\ 12.6) and coding (HumanEval: 42.1 vs.\ 21.3), highlighting the superior training efficiency and generalization of the AR paradigm under identical compute budgets.

\textbf{Second}, our proposed conversion strategy successfully preserves the strength of the AR backbone. AR-BD-2B-Chat (SDAR) achieves performance largely on par with AR-2B-Chat, even slightly surpassing it on knowledge-intensive tasks such as MMLU (+2.2) and GPQA (+1.9). The small trade-offs observed on MATH (-3.1) and HumanEval (-2.1) confirm that the conversion incurs negligible quality loss while enabling parallel decoding. This validates our central claim: AR models can be efficiently adapted into the Block Diffusion paradigm at low cost.

\textbf{Third}, converting MDLMs into Block Diffusion also yields significant gains. MDLM-BD-2B-Chat consistently outperforms its own MDLM baseline, with notable improvements on reasoning and coding benchmarks (e.g., MATH nearly doubles from 12.6 to 23.3). However, even after this improvement, MDLM-BD still lags behind AR-BD in most domains, reaffirming that AR is the stronger architectural starting point for Block Diffusion adaptation.

In summary, the results demonstrate that (i) AR baselines are inherently more effective than MDLMs, (ii) AR-BD maintains the performance of AR while adding parallel decoding, and (iii) MDLM-BD benefits from the conversion but cannot close the fundamental gap with AR-based models.

\section{Scaling SDAR: Principles and Practice}\label{part2}
\begin{tcolorbox}[takeawaysbox]
(1) Low-cost AR to SDAR adaptation works for any modern AR base, delivering \textbf{on-par performance with broad applicability.}

(2) Larger SDAR models tolerate bigger blocks and looser decoding thresholds, enabling \textbf{higher parallel efficiency without sacrificing performance.}

(3) In SDAR, larger models and higher confidence not only improve quality but also drive faster decoding, \textbf{making accuracy the engine of efficiency.}

\end{tcolorbox}
In Section~\ref{part1}, we conducted a rigorous and fair comparison that established the foundational viability of our {SDAR} framework. However, those initial experiments were performed on a relatively small scale, utilizing a 1.7B parameter model trained on 1T tokens. To thoroughly investigate the scalability of our approach and develop production-ready models, this section extends our study to a much larger and more capable family of models. 

To mature {SDAR} into a production-ready paradigm, our investigation seeks to replace empirical guesswork with a principled understanding of its scaling behavior. This requires answering two fundamental, sequential questions:
\begin{itemize}
    \item \textbf{Intrinsic Efficiency of the SDAR Paradigm}: What is the intrinsic inference speedup of {SDAR} in downstream tasks, free of any external optimizations? This establishes the core efficiency we seek to scale.
    \item \textbf{The Interplay of Model Scale and Block Size}: How does this efficiency depend on the interplay between model size and `blocksize`? By modeling this relationship, we aim to derive a predictive scaling law, enabling practitioners to configure larger models optimally and efficiently, thus providing a critical roadmap for future work.
\end{itemize}

\subsection{Scaling Performance of the Qwen3 Family on Downstream Benchmarks}\label{part:2_1}
\subsubsection{Experimental Setup}\label{eval_setup}
\paragraph{Training Setup}
We investigate scaling properties using the Qwen3 series~\citep{yang2025qwen3} as our base models, starting from the Qwen3-1.7B, 4B, 8B, and 30B-A3B base models. To adapt these models for downstream tasks, we adopt a two-stage training strategy:
\begin{enumerate}
    \item \textbf{Continued Pre-training (CPT):} We first continue pre-training each base model on a 50B-token subset randomly sampled from the 1T-token corpus described in Section~\ref{part1}, using a context length of 4096 with packed sequences.
    \item \textbf{Supervised Fine-tuning (SFT):} Following CPT, the models are supervised fine-tuned on a high-quality 4B-token instruction-following dataset.
\end{enumerate}
This procedure is applied to generate our main models, the \textbf{SDAR-Chat} series (1.7B, 4B, 8B, 30B-A3B). To establish strong baselines, we also apply the identical CPT and SFT pipeline to produce standard autoregressive counterparts, termed \textbf{AR-Chat} series.

\paragraph{Evaluation Setup}

We conduct a comprehensive evaluation of our models on a suite of downstream benchmarks designed to assess key capabilities in reasoning, mathematics, code generation, and instruction following. The evaluation datasets are organized as follows:
\begin{itemize}
    \item \textbf{Reasoning \& Knowledge:} , MMMLU-lite~\citep{hendryckstest2021} (0-shot), TriviaQA~\citep{joshi-etal-2017-triviaqa} (1-shot), MMLU~\citep{hendryckstest2021} (5-shot), CMMLU~\citep{li2023cmmlu} (0-shot), MMLU-Pro~\citep{wang2024mmlu} (0-shot, CoT), and GPQA-diamond~\citep{rein2024gpqa} (0-shot).
    \item \textbf{Mathematics:} GSM8K~\citep{cobbe2021gsm8k} (0-shot, CoT), MATH-500~\citep{lightman2023lets} (0-shot, CoT), MathBench (0-shot, CoT)~\citep{liu2024mathbench}, and the challenging competition-level benchmarks AIME-2024, AIME-2025~\citep{aime25} and LiveMathBench-Hard (LMB-Hard)~\citep{liu2024your}.
    \item \textbf{Code Generation:} HumanEval~\citep{chen2021codex}, HumanEval-X~\citep{zheng2023codegeex}, MBPP~\citep{austin2021program}, and LiveCodeBench (LCB)~\citep{jain2024livecodebench}, all evaluated in a zero-shot setting.
    \item \textbf{Instruction Following:} IFEval~\citep{zhou2023instruction} (0-shot).
\end{itemize}

For our evaluation protocol, we employ greedy static decoding for the SDAR-Chat models, with both the block length and the number of denoising steps set to 4. To ensure a fair comparison, all autoregressive models, including our AR-Chat and external baselines, are evaluated using standard greedy decoding. Our primary baselines are the original Qwen3-Base series, with performance metrics cited directly from the Qwen3 Technical Report~\citep{yang2025qwen3}. To position our work in a broader context, we also include results for LLaDA and Dream, as reported in~\citep{nie2025large, ye2025dream}.

\subsubsection{Summary of Evaluation Results}
Our comprehensive evaluation reveals several key insights about the scaling behavior and performance characteristics of SDAR models compared to their autoregressive counterparts and other diffusion-based approaches.

\begin{table*}[ht]
\centering
\small
\caption{Comprehensive performance comparison of AR and SDAR models. $^\dag$ indicates that the models are derived from the Qwen3 Base models by performing the identical CPT and SFT. For each benchmark, the best result within a given scale (e.g., 1.7B, 30B) is shown in \textbf{bold}. The colored superscript for SDAR models indicates the performance difference relative to the corresponding AR-Chat baseline.}~\label{tab:comprehensive_results_superscript}
\resizebox{\linewidth}{!}{
\begin{tabular}{l  c c c c  c c c c}
\toprule
\textbf{Model} & \multicolumn{4}{c}{\textbf{AR-Chat Model}} & \multicolumn{4}{c}{\textbf{SDAR-Chat Model}} \\
\cmidrule(r){2-5} \cmidrule(l){6-9}
{scale} & {Qwen3-1.7B$^\dag$} & {Qwen3-4B$^\dag$} & {Qwen3-8B$^\dag$} & {Qwen3-30BA3B$^\dag$} & {SDAR-1.7B} & {SDAR-4B} & {SDAR-8B} & {SDAR-30BA3B} \\
\midrule
\multicolumn{9}{c}{{Reasoning, Knowledge \& Instruction Following}} \\
\midrule
ARC-C            & 81.0 & 89.5 & 91.9 & 93.9   & \valdelta{\textbf{85.4}}{+4.4} & \valdelta{\textbf{90.5}}{+1.0} & \valdelta{\textbf{91.9}}{0.0}  & \valdelta{93.2}{-0.7} \\
TriviaQA         & 45.0 & 57.9 & 68.2 & 66.5   & \valdelta{42.6}{-2.4}          & \valdelta{57.2}{-0.7}          & \valdelta{\textbf{69.2}}{+1.0} & \valdelta{\textbf{75.9}}{+9.4} \\
MMLU             & 63.8 & 74.8 & 77.5 & 82.2   & \valdelta{62.9}{-0.9}          & \valdelta{\textbf{74.9}}{+0.1} & \valdelta{\textbf{78.6}}{+1.1} & \valdelta{\textbf{82.8}}{+0.6} \\
MMLU-Pro         & 39.0 & 52.8 & 56.6 & 63.8   & \valdelta{37.0}{-2.0}          & \valdelta{50.9}{-1.9}          & \valdelta{\textbf{56.9}}{+0.3} & \valdelta{61.5}{-2.3} \\
GPQA-diamond     & 28.6 & 31.4 & 37.0 & 37.3   & \valdelta{\textbf{29.8}}{+1.2} & \valdelta{\textbf{33.0}}{+1.6} & \valdelta{\textbf{40.2}}{+3.2} & \valdelta{36.7}{-0.6} \\
IFEval           & 43.3 & 58.4 & 60.8 & 57.7   & \valdelta{\textbf{43.4}}{+0.1} & \valdelta{56.6}{-1.8}          & \valdelta{\textbf{61.4}}{+0.6} & \valdelta{\textbf{60.6}}{+2.9} \\
\midrule
\multicolumn{9}{c}{{Mathematics}} \\
\midrule
GSM8K            & 81.1            & 90.7           & 92.8           & 92.7   & \valdelta{80.1}{-1.0}          & \valdelta{89.9}{-0.8}                     & \valdelta{91.3}{-1.5}            & \valdelta{91.4}{-1.3} \\
Math500          & 62.0             & 74.1           & 78.4           & 76.8   & \valdelta{\textbf{63.2}}{+1.2} & \valdelta{72.8}{-1.3}                     & \valdelta{\textbf{78.6}}{+0.2}   & \valdelta{\textbf{77.8}}{+1.0} \\
MathBench        & 60.5             & 70.6           & 73.5           & 78.4   & \valdelta{\textbf{63.6}}{+3.1} & \valdelta{\textbf{74.7}}{+4.1}            & \valdelta{\textbf{76.9}}{+3.4}   & \valdelta{\textbf{79.3}}{+0.9} \\
AIME-24          & \phantom{0}7.1   & 12.9           & 10.0           & 15.4   & \valdelta{\textbf{10.0}}{+2.9} & \valdelta{\textbf{10.0}}{-2.9}            & \valdelta{10.0}{0.0}             & \valdelta{\textbf{16.7}}{+1.3} \\
AIME-25          & \phantom{0}3.3   & \phantom{0}5.0 & \phantom{0}7.5 & 10.8   & \valdelta{\phantom{0}2.1}{-5.9}& \valdelta{\textbf{\phantom{0}7.5}}{+2.5}  & \valdelta{\textbf{10.0}}{+2.5}   & \valdelta{10.8}{+0.0} \\
LMB-Hard         & \phantom{0}9.2   & 11.6           & 15.7           & 15.5   & \valdelta{\phantom{0}6.6}{-2.6}& \valdelta{\phantom{0}6.9}{-4.7}           & \valdelta{\phantom{0}8.9}{-6.8}  & \valdelta{13.7}{-1.8} \\
\midrule
\multicolumn{9}{c}{{Code Generation}} \\
\midrule
HumanEval        & 65.9             & 73.4 & 80.7 & 84.8            & \valdelta{61.6}{-4.3}          & \valdelta{72.8}{-0.6}            & \valdelta{78.7}{-2.0}          & \valdelta{\textbf{87.2}}{+2.4} \\
MBPP             & 61.9             & 67.1 & 75.1 & 75.1            & \valdelta{61.1}{-0.8}          & \valdelta{65.4}{-1.7}            & \valdelta{72.0}{-3.1}          & \valdelta{\textbf{71.6}}{+3.5} \\
HumanEval-X      & 47.0             & 60.9 & 63.5 & 63.5            & \valdelta{45.2}{-1.8}          & \valdelta{\textbf{62.3}}{+1.4}   & \valdelta{\textbf{64.9}}{+1.4} & \valdelta{\textbf{66.3}}{+2.8} \\
LCB-v6           & \phantom{0}8.6   & 10.3 & 20.5 & 23.4            & \valdelta{\phantom{0}5.7}{-2.9}& \valdelta{\textbf{13.1}}{+2.8}   & \valdelta{16.6}{-3.9}          & \valdelta{21.7}{-1.7} \\
\midrule
\multicolumn{9}{c}{Language} \\
\midrule
MMMLU-lite       & 36.7 & 44.4 & 55.2 & 52.9       & \valdelta{40.9}{+4.2} & \valdelta{\textbf{50.7}}{+6.3}  & \valdelta{\textbf{55.3}}{+0.1}   &\valdelta{57.2}{+4.3}\\
CMMLU            & 60.7 & 72.6 & 76.2 & 82.0       & \valdelta{60.2}{-0.5} & \valdelta{71.3}{-1.3}           & \valdelta{75.7}{-0.5}            & \valdelta{81.0}{-1.0} \\
\bottomrule
\end{tabular}%
}
\end{table*}

\paragraph{Scaling Without Compromise}
As detailed in Table~\ref{tab:comprehensive_results_superscript}, the {SDAR} paradigm demonstrates robust scaling properties analogous to traditional AR models. As we increase model size from 1.7B to 30B parameters, performance consistently improves across all benchmark categories, with MMLU scores scaling from 62.9\% to 82.8\%, for instance. This confirms that the benefits of scale are effectively preserved.
Crucially, this scaling does not come at the cost of performance. At the 30B scale, {SDAR-Chat} achieves parity or surpasses its {AR-Chat} counterpart on 11 out of 18 benchmarks. This advantage is particularly pronounced in complex reasoning domains. For instance, {SDAR} models show a notable aptitude for structured generation, outperforming the AR model on key code generation tasks such as HumanEval (+2.4\%) and HumanEval-X (+2.8\%). They also exhibit consistently strong performance on mathematical benchmarks like Math500 and MathBench. These results establish that the {SDAR} framework not only scales effectively but also preserves, and in some cases enhances, the core reasoning capabilities of the base models.

\paragraph{Positioning SDAR as a State-of-the-Art Paradigm}
In Table~\ref{tab:sdar_vs_qwen3}, we further contextualize these findings by comparing {SDAR} with the original Qwen3-Base models and other prominent non-autoregressive architectures like LLaDA~\citep{nie2025large} and Dream~\citep{ye2025dream}. Our {SDAR} models substantially outperform these prior diffusion-based approaches across all key benchmarks. For example, our {SDAR-8B} model achieves 78.6\% on MMLU, a significant leap over LLaDA-8B (65.9\%) and Dream-7B (69.5\%). This substantial performance gap highlights that {SDAR} represents a major advancement for non-autoregressive generation, making it truly competitive with top-tier AR models for the first time.
Furthermore, the {SDAR-Chat} models demonstrate enhanced instruction-following capabilities over their {AR-Chat} counterparts (e.g., a +2.9\% gain on IFEval at the 30B scale), suggesting the block-wise approach may better reinforce alignment.
Taken together, these results validate that our two-stage training strategy provides a robust foundation and positions {SDAR} as a leading paradigm, not only advancing beyond previous diffusion models but also establishing itself as a powerful and capable competitor to autoregressive generation.

\paragraph{Highly Efficient Adaptation} 
A key advantage of our approach lies in its extremely low adaptation cost. Unlike DiffuLLaMA, which relies on 65B tokens for conversion, or Dream, which requires 580B tokens yet still suffers from substantial degradation relative to their AR bases (LLaMA and Qwen, respectively), {SDAR} achieves strong performance with only 50B tokens of continued pretraining. This efficiency stems from the fact that AR pretraining provides dense supervision, so the subsequent block-diffusion adaptation requires only a short training phase to align objectives. We further validate this by experimenting with smaller adaptation budgets of 20B, 30B, and 40B tokens. While these settings also yield functional SDAR models, they result in slightly lower downstream task performance and reduce flexibility for scaling block sizes during subsequent SFT. Importantly, even at these lower budgets, the degradation is far less severe than observed in prior diffusion-based conversions. Moreover, our adaptation relies on relatively low-quality open-source corpora, in contrast to the high-quality synthetic 6T data used in the final stage of Qwen3 pretraining. We therefore believe that with better data quality and distributional alignment, the adaptation phase could operate with an even smaller budget without sacrificing performance.

\begin{table*}[ht]
\centering
\small
\caption{Performance comparison between SDAR and AR models at different scales. Best results within each scale are in {bold}.}
\resizebox{0.95\linewidth}{!}{ 
\begin{tabular}{l l c c cc c c}
\toprule
\textbf{Scale} & \textbf{Model} & \textbf{MMLU} & \textbf{GSM8K} & \textbf{Math500}  & \textbf{HumanEval} & \textbf{MBPP} & \textbf{IFEval} \\
\midrule
\multicolumn{8}{c}{{AR Models}} \\
\midrule
1.7B & AR-Chat  & {63.8} & {81.1} & 62.0  & {65.9} & {61.9} & 43.3 \\
     & Qwen3-Base    & 62.6 & 75.4 & 43.5   & --   & 55.4 & -- \\
30BA3B  & AR-Chat   & 82.2 & 92.7 & 76.8  & 84.8 & 75.1 & 57.7 \\
     & Qwen3-Base     & 81.4 & 91.8 & 59.0    & --   & 74.4 & -- \\
\midrule
\multicolumn{8}{c}{Diffusion Models} \\
\midrule
8B   & LLaDA           & 65.9   & 78.6 & 37.3      & 47.6 & 34.2 & 59.9 \\
7B   & Dream           & 69.5   & 81.0 & 38.7     & 55.5 & 58.8 & \textbf{62.5} \\
\midrule
\multicolumn{8}{c}{{SDAR Models}} \\
\midrule
1.7B & \multirow{4}{*}{{SDAR-Chat}}
     & 62.9 
     & 80.1 
     & 63.2 
     & 61.6 
     & 61.1 
     & 43.4  \\
4B   &         & 74.9 & 89.9 & 72.8  & 72.0 & 65.4 & 56.6 \\
8B   &         & 78.6 & 91.3 & 78.6  & 78.7 & 72.0 & 61.4 \\
30BA3B  & 
     & 82.8 
     & 91.4 
     & 77.8 
     & 87.2 
     & 71.6 
     & 60.6  \\
\bottomrule
\end{tabular}}
\label{tab:sdar_vs_qwen3}
\end{table*}

\subsection{Scaling Dynamics: A Trade-off Analysis of Performance, Efficiency, and Model Dimensions}

To systematically chart the performance landscape of the {SDAR} framework, we conduct a multi-dimensional analysis across model scale, architectural hyperparameters, and inference strategies. This factorial experimental design allows us to deconstruct the intricate trade-offs between generative performance and computational efficiency. The resulting analysis provides a principled, empirical guide for configuring and deploying {SDAR} models to meet specific operational requirements.

\subsubsection{Experimental Setup}
Our analysis systematically varies three primary factors:

\begin{itemize}
    \item \textbf{Model Scale:} We evaluate all models in the {SDAR-Chat} family: 1.7B, 4B, 8B, and 30B-A3B.
    
    \item \textbf{Block Size ($B$):} We sweep the block size, a core hyperparameter of {SDAR}, across a range of values from 4 to 64.

    \item \textbf{Decoding Strategy:} We investigate two distinct decoding regimes:
    \begin{itemize}
        \item \textbf{Static Decoding:} In this mode, the generation of each block involves a full sequence of $B$ iterative denoising steps, where $B$ is the model's architectural block size. This exhaustive refinement process maximizes output quality, thereby establishing the \textbf{performance ceiling} for the given architecture. All dynamic, early-exit strategies are thus evaluated as trade-offs against this quality benchmark.
        
        \item \textbf{Dynamic Decoding:} To enable a finer-grained trade-off, this mode dynamically adjusts the number of tokens generated per step, $k$ (where $1 \le k \le B$), based on a confidence threshold $\tau$. The number of denoising steps equals the chosen block size for that step. We evaluate a range of thresholds, $\tau \in \{0.80, 0.85, 0.90, 0.95\}$, to explore the spectrum between aggressive and conservative parallel generation.
    \end{itemize}
\end{itemize}

\paragraph{Evaluation Metrics.} We assess the models along two axes:

\begin{itemize}
    \item \textbf{Performance:} Model performance is quantified using the comprehensive benchmark suite detailed in Section~\ref{eval_setup}.
    
    \item \textbf{Efficiency:} We measure computational efficiency via the \textbf{Effective Tokens Per Forward Pass (TPF)}, defined as:
    \begin{equation*}
        \text{TPF} = \frac{\text{Total Generated Tokens}}{\text{Total Forward Passes}}
    \end{equation*}
    This hardware-agnostic metric directly quantifies the algorithmic speedup of parallel decoding by measuring the average token throughput per inference step. The relative speedup is then benchmarked against a standard autoregressive (AR) model, for which TPF is axiomatically~1.
\end{itemize}

\subsubsection{Performance Landscape Across Model Scale and Block Size}

\begin{figure}[!htbp]
    \centering
    \includegraphics[width=0.99\textwidth]{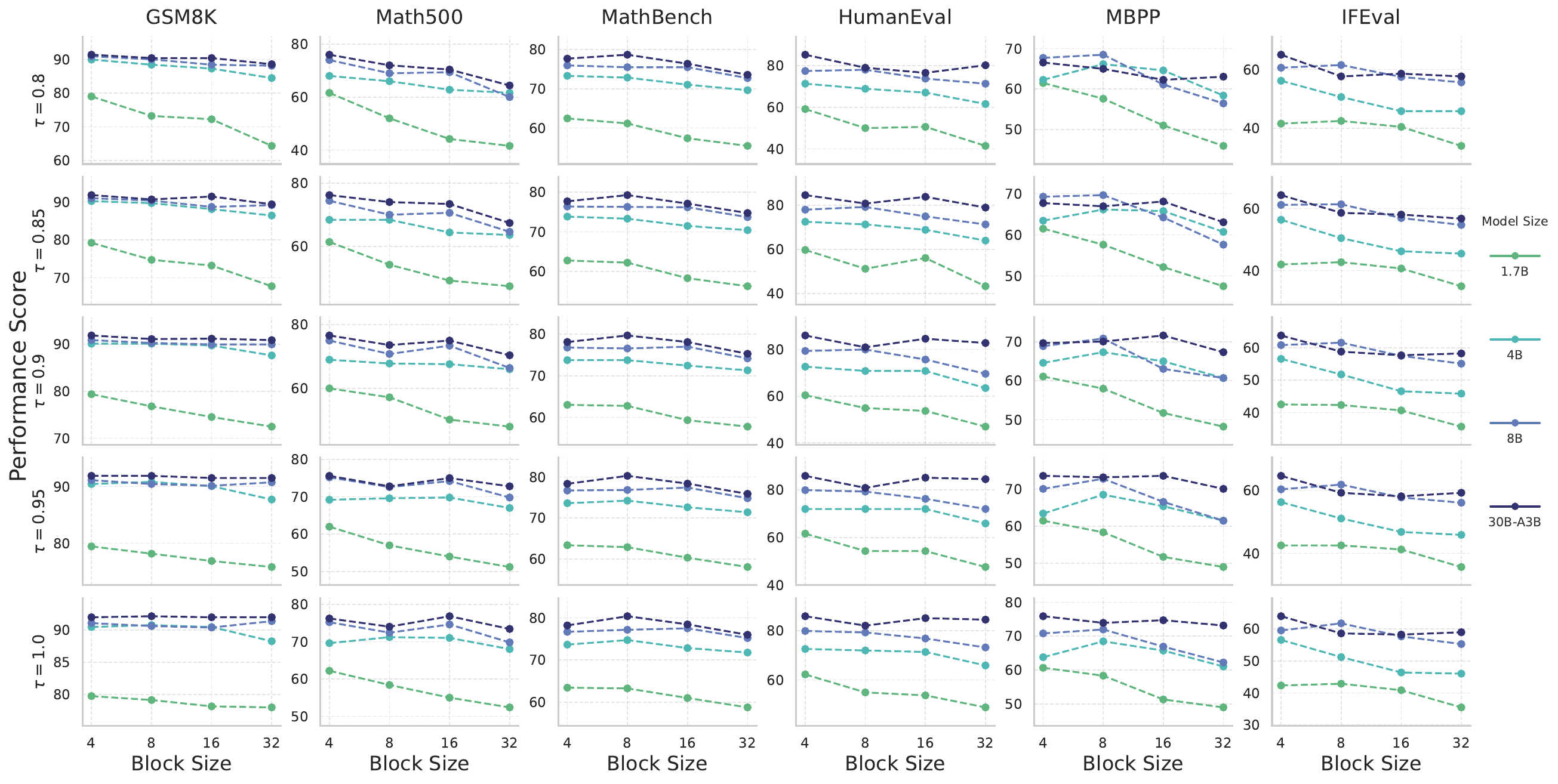}
    \caption{Performance on various benchmarks as a function of architectural block size ($B$) across different model scales and various confidence thresholds ($\tau$).}
    \label{fig:Performance_blksz}
\end{figure}

\begin{figure}[!htbp]
    \centering
    \includegraphics[width=0.99\textwidth]{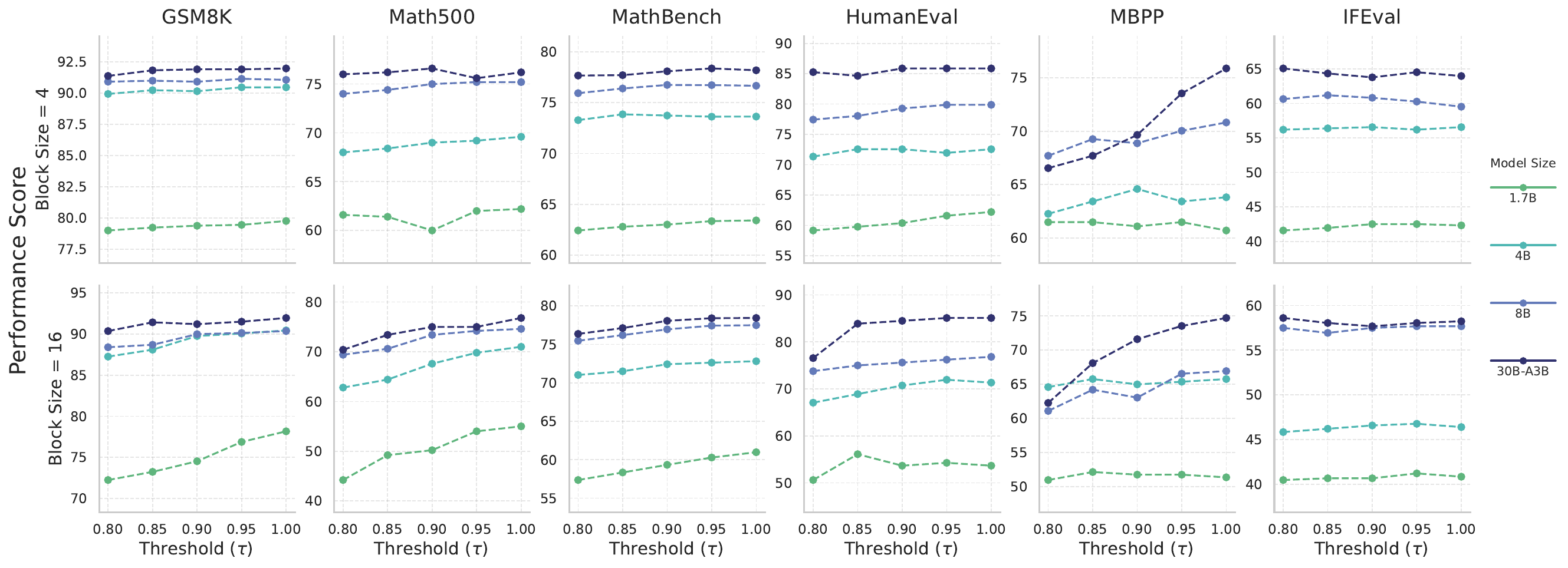}
    \caption{Performance as a function of the dynamic decoding confidence threshold ($\tau$) for block sizes $B=4$ and $B=16$.}
    \label{fig:Performance_vs_tau_bs4_16}
\end{figure}

Figure~\ref{fig:Performance_blksz} and Figure~\ref{fig:Performance_vs_tau_bs4_16} chart the generative performance of {SDAR} models as a function of their architectural block size ($B$) and the dynamic decoding confidence threshold ($\tau$). Our analysis reveals several critical scaling dynamics.

First, we observe a clear interaction between model scale and robustness to parallelization. Smaller models (1.7B, 4B) exhibit significant sensitivity to increases in block size, with performance degradation becoming pronounced for $B>4$. In stark contrast, larger models (8B, 30B-A3B) demonstrate remarkable resilience, maintaining stable performance across a much wider range of block sizes. This unveils a \textbf{virtuous cycle of scaling}: larger, more capable models tolerate larger block sizes without a commensurate performance drop, directly enabling them to achieve substantially greater parallelization and efficiency. This synergy, where improved capability naturally translates into improved computational throughput, underscores the profound potential of the {SDAR} framework at scale.

Furthermore, the relationship between performance and block size is not purely monotonic. While small block sizes ($B=4$) yield performance nearly indistinguishable from their autoregressive counterparts, and larger block sizes ($B=32$) often introduce task-dependent trade-offs, we identify an interesting inflection point. For intermediate block sizes ($B \in \{8, 16\}$), performance is not only maintained but, on certain complex reasoning tasks such as GSM8k, MathBench, and HumanEval, can even surpass that of the smaller-block configurations. This suggests that a moderately larger generation context can, in some cases, provide a beneficial inductive bias.

The analysis of the confidence threshold $\tau$ (Figure~\ref{fig:Performance_vs_tau_bs4_16}) reinforces these scaling laws. As expected, performance consistently improves as $\tau$ approaches 1.0 (transitioning from dynamic to static decoding). More importantly, larger models display a higher tolerance for more aggressive (i.e., lower $\tau$) decoding strategies. For the 30B-A3B model, performance remains nearly saturated for $\tau \ge 0.90$ across most tasks and block sizes. This dual robustness—to both larger architectural block sizes and more lenient decoding thresholds—is a key finding, indicating that the benefits of scale compound to enable more aggressive parallelization without sacrificing quality. Notably, MBPP shows a stronger sensitivity to $\tau$ compared to other benchmarks because it targets base model evaluation, for which our models after SFT are less suited, resulting in larger performance fluctuations as $\tau$ changes.

\subsubsection{Efficiency Dynamics and the Performance-Speedup Interplay}
\begin{figure}[!htbp]
    \centering
    \includegraphics[width=0.99\textwidth]{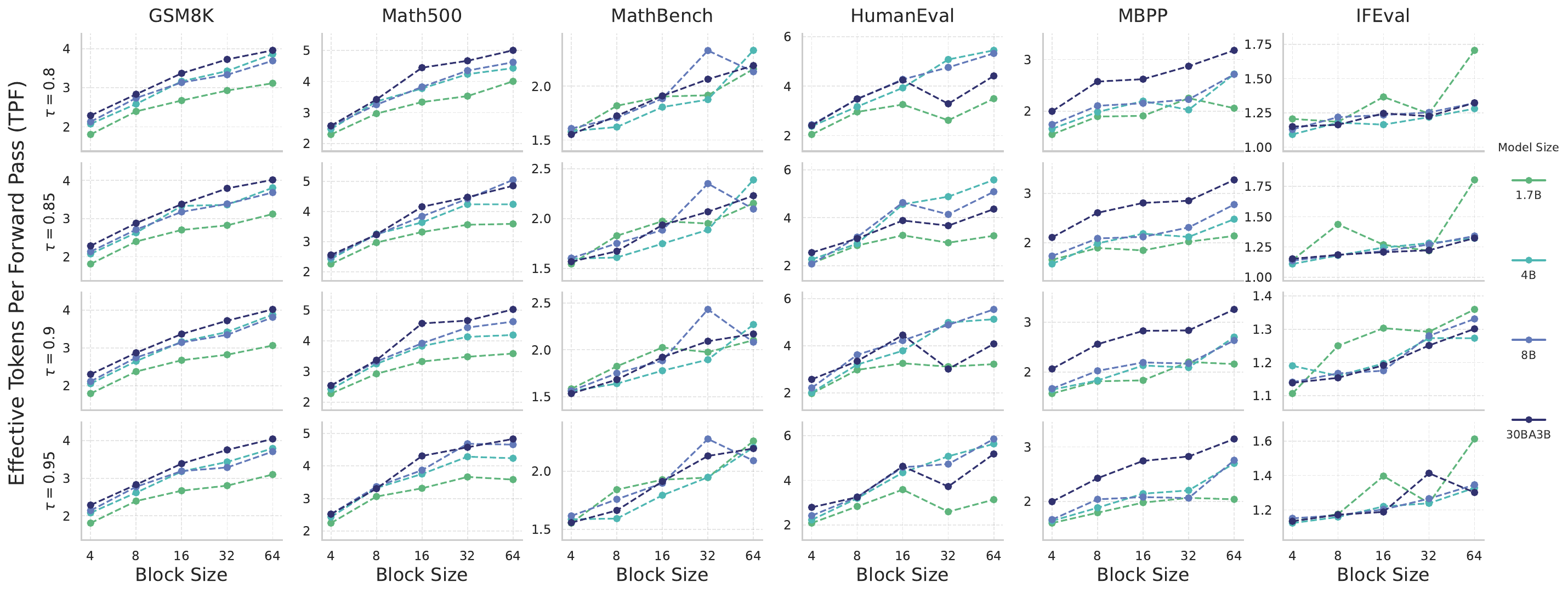}
    \caption{Algorithmic speedup, measured in Effective Tokens Per Forward Pass (TPF), as a function of architectural block size ($B$) across different model scales and various confidence thresholds ($\tau$).}
    \label{fig:Speed_blksz}
\end{figure}

We now turn to the efficiency axis, quantified by Effective Tokens Per Forward Pass (TPF), which measures the algorithmic speedup over standard autoregressive decoding. 

Figure~\ref{fig:Speed_blksz} presents a clear and encouraging result: TPF scales predictably and monotonically with the block size $B$. This relationship holds true across all model scales, tasks, and decoding thresholds, confirming that a larger architectural block size serves as a reliable lever for increasing computational throughput. A second, more profound trend also emerges: \textbf{larger models consistently achieve higher TPF}. This phenomenon is attributable to the lower predictive entropy and higher confidence inherent in more capable models. For a given block, a larger model is more likely to make confident predictions that surpass the threshold $\tau$, thus generating more tokens in parallel. This points to a promising avenue for future optimization: \textbf{techniques that reduce a model's predictive entropy, such as knowledge distillation~\citep{song2025seed} or reinforcement learning-induced "entropy collapse"~\citep{cui2025entropy}, could directly translate into greater decoding efficiency.}

\begin{figure}[!htbp]
    \centering
    \includegraphics[width=0.99\textwidth]{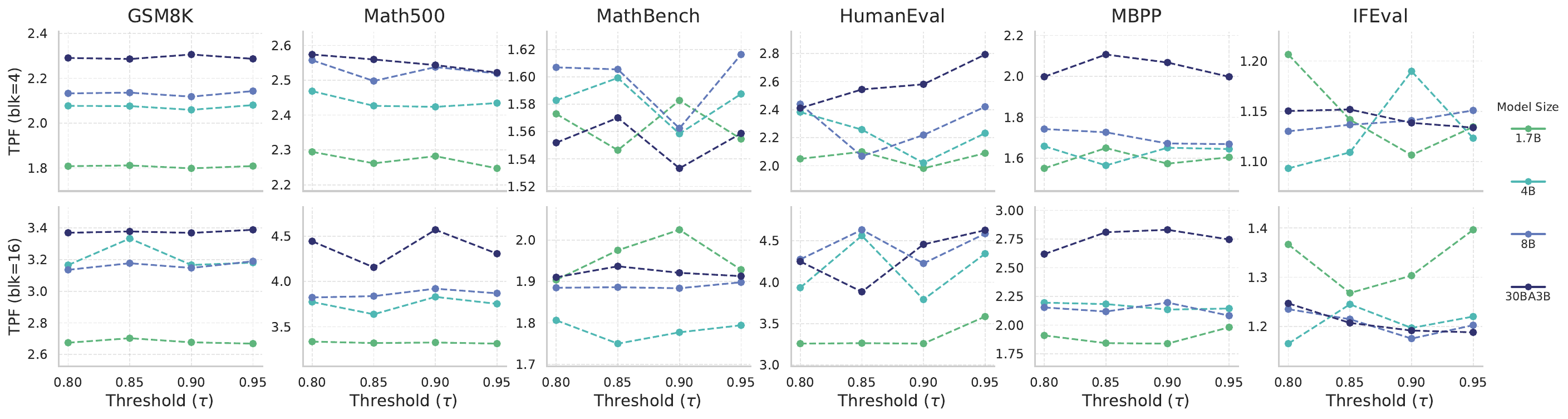}
    \caption{Algorithmic speedup (TPF) as a function of the dynamic decoding confidence threshold ($\tau$) for block sizes $B=4$ and $B=16$.}
    \label{fig:Speed_vs_tau_blk4_16}
\end{figure}
However, the interaction between TPF and the confidence threshold $\tau$, depicted in Figure~\ref{fig:Speed_vs_tau_blk4_16}, reveals a more complex dynamic. Contrary to the naive intuition that a lower threshold would always yield higher TPF, our results show a non-monotonic and task-dependent relationship. This is because decoding is a sequential Markov process where the quality of one generated block conditions the next. An overly aggressive (low) threshold may lead to the acceptance of a low-quality, high-entropy block. This error propagation increases the uncertainty for subsequent steps, lowering their predictive confidence and, consequently, reducing the number of tokens generated in parallel.

This exposes our central finding on the trade-off calculus: \textbf{in the {SDAR} framework, performance and efficiency are not independent axes but are deeply intertwined.} Striving for higher generative quality—by using a more conservative threshold or a more capable model—can, paradoxically, lead to improved computational efficiency by maintaining a low-entropy, high-confidence state throughout the decoding process. This reframes the role of model quality entirely. \textbf{In the {SDAR} framework, high confidence and low entropy are not outcomes to be balanced against speed, but are themselves the very engine of acceleration.}

\subsubsection{Deployment-Ready Efficiency Analysis on Industrial Inference Engines}

\begin{figure}[!htbp]
    \centering
    \includegraphics[width=0.99\textwidth]{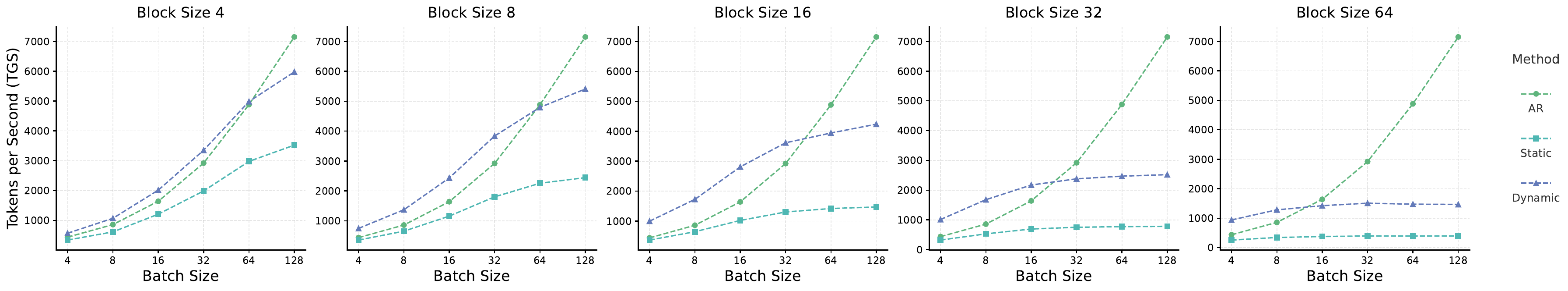}
    \caption{Throughput scaling characteristics (tokens per second) across varying batch sizes for autoregressive (AR) decoding versus SDAR with different block sizes ($B \in \{4, 8, 16, 32, 64\}$). Results shown for SDAR-8B-Chat under both static and dynamic decoding regimes.}
    \label{fig:tgs_comparison}
\end{figure}

While theoretical speedup metrics provide valuable insights into algorithmic efficiency, the translation of these gains into real-world deployment scenarios remains a critical question. We therefore conducted a comprehensive analysis of SDAR's performance on industrial-grade inference infrastructure, addressing two fundamental questions:
\begin{enumerate}
    \item What absolute throughput can SDAR achieve on production-ready inference engines?
    \item How effectively does the theoretical TPF translate to real-world inference speed?
\end{enumerate}

\paragraph{Experimental Setup} Our experimental setup employed the LMDeploy inference engine on a single NVIDIA H200 GPU, using SDAR-Chat-8B models with block sizes ranging from 4 to 64. All dynamic decoding experiments utilized a confidence threshold of $\tau=0.9$, which our earlier analysis showed preserves quality parity with standard autoregressive generation. We measured throughput across batch sizes from 4 to 512 on the Math-500 benchmark.

\paragraph{Analysis Framework} To formalize our analysis, we define: $t_f$ (time per forward pass), $\alpha$ (batch size), $\beta$ (block size), and $\gamma$ (average forward passes per block). For standard autoregressive decoding:
\begin{align}
    \text{TGS}_{\text{AR}} &= \frac{\alpha}{t_f} \\
    t_f &= \frac{\alpha}{\text{TGS}_{\text{AR}}}
\end{align}

With our inference engine requiring an additional forward pass for KV cache computation, static SDAR decoding throughput becomes:
\begin{align}
    \text{TGS}_{\text{static}} &= \frac{\alpha\beta}{(\beta+1)t_f} \\
    t_f &= \frac{\alpha\beta}{(\beta+1)\text{TGS}_{\text{static}}}
\end{align}

For dynamic decoding with confidence-based forward passes:
\begin{align}\label{cal_tpf}
    \text{TGS}_{\text{dynamic}} &= \frac{\alpha\beta}{(\gamma+1)t_f} \\
    t_f &= \frac{\alpha\beta}{(\gamma+1)\text{TGS}_{\text{dynamic}}} \\
    \text{TPF} &= \frac{\beta}{\gamma} = \frac{1}{\frac{\alpha}{\text{TGS}_{\text{dynamic}} \times t_f}-\frac{1}{\beta}}
\end{align}

\paragraph{Analysis}
\begin{figure}[t]
    \centering
    \includegraphics[width=0.99\textwidth]{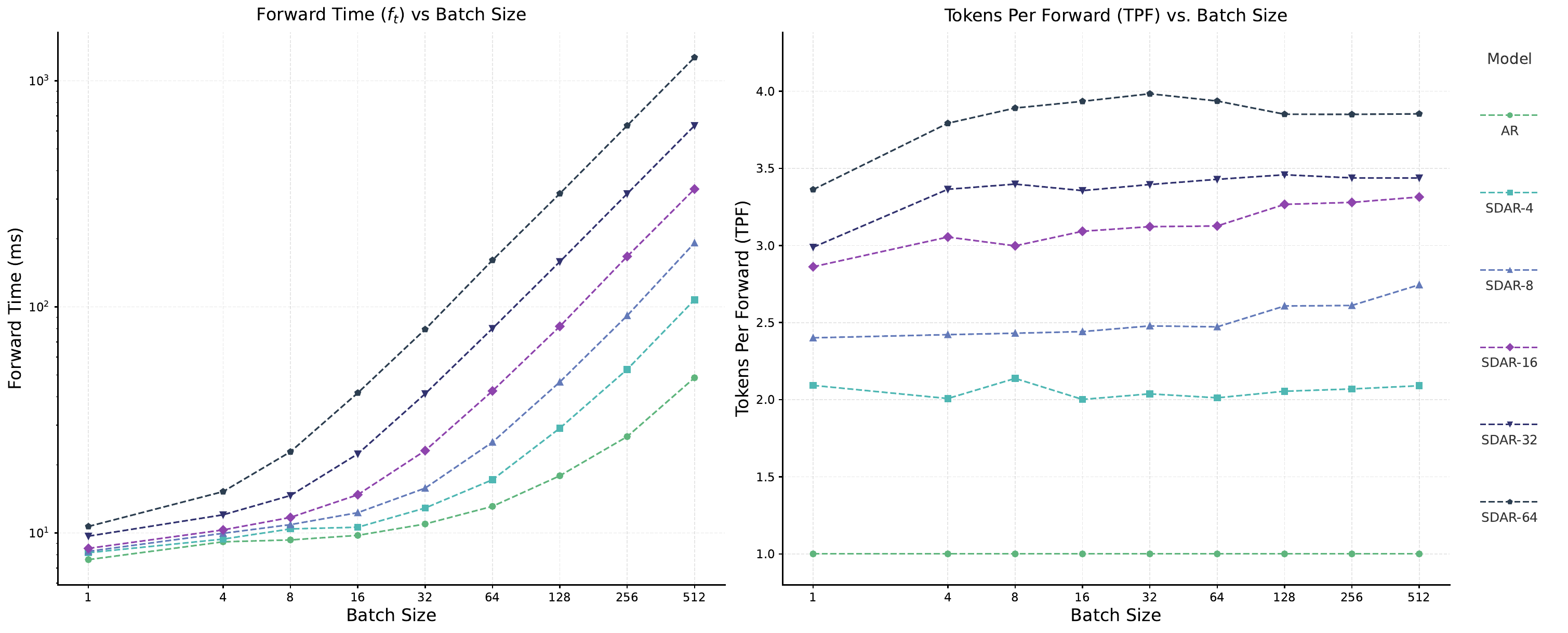}
    \caption{\textbf{Left:} Single forward pass time ($t_f$) as a function of batch size for AR and SDAR models with varying block sizes. The linear relationship at higher batch sizes indicates compute-bound operation. \textbf{Right:} Realized Tokens Per Forward Pass (TPF) calculated from actual inference engine throughput, demonstrating the practical manifestation of theoretical speedup across different block sizes.}
    \label{fig:ft_and_tpf_comparison_optimized}
\end{figure}

Figure~\ref{fig:tgs_comparison} reveals several critical insights about SDAR's real-world performance characteristics. First, our SDAR models achieve impressive peak throughput under dynamic decoding, with maximum tokens generated per second (TGS) of 6600, 5510, 4236, 2520, and 1509 for block sizes 4, 8, 16, 32, and 64, respectively. Second, we observe a clear batch size dependency: SDAR models significantly outperform autoregressive decoding at smaller batch sizes, but this advantage diminishes as batch size increases. This phenomenon stems from SDAR's higher computational intensity per forward pass, which causes it to reach compute-bound operation earlier than standard autoregressive models.

This batch size sensitivity exhibits a clear relationship with block size—larger blocks provide greater acceleration advantages in low-batch regimes but encounter compute limitations more rapidly as batch size increases. This finding has profound implications for deployment strategies: \textbf{SDAR provides the greatest relative advantage in latency-sensitive, low-batch inference scenarios}, precisely the operating conditions most relevant for real-time applications and interactive services.

Examining Figure~\ref{fig:ft_and_tpf_comparison_optimized}, we gain further insight into the underlying mechanics of this performance profile. The left panel demonstrates that forward pass time increases gradually with batch size until reaching a compute-bound threshold, after which the relationship becomes strictly linear with consistent slope across models. This inflection point represents the maximum effective parallelism the hardware can support for each model configuration.

The right panel of Figure~\ref{fig:ft_and_tpf_comparison_optimized} presents the realized TPF calculated using Equation~\ref{cal_tpf}, which closely aligns with our theoretical measurements in Figure~\ref{fig:Speed_blksz}. This correspondence validates that the algorithmic speedup of SDAR successfully translates to practical deployment environments, though the absolute throughput advantage depends on the specific operating regime.

These findings suggest several optimization pathways for future inference engine implementations. First, eliminating the separate KV cache computation by integrating it with subsequent decoding steps could further improve throughput. Second, specialized kernel optimizations for the unique computation patterns of SDAR could extend the batch size range where SDAR maintains its advantage. Finally, the clear relationship between block size, batch size, and throughput enables deployment-time configuration optimization based on specific workload characteristics and latency requirements.
\section{Reasoning potential of SDAR}
\begin{tcolorbox}[takeawaysbox]
(1) \textbf{Reasoning preserved.} SDAR adaptation retains and easily restores long CoT reasoning from strong AR reasoning models.

(2) \textbf{Domain shift for free.} SDAR enables effortless domain transfer during adaptation, excelling on tasks benefiting from local bidirectional context.

(3) \textbf{Scaling synergy.} Simple test-time scaling brings large gains, suggesting strong potential for further improvement via RL.

\end{tcolorbox}

Over just the past year, the reasoning capabilities of large language models have emerged as a central focus of the field. Yet these advances typically come with a steep cost: test-time scaling methods place a heavy burden on inference. A core motivation of our study is the hypothesis that SDAR’s parallel decoding capability can mitigate this challenge, offering a more efficient pathway to test-time scaling. To examine this potential, we adapt strong autoregressive (AR) baselines into the SDAR paradigm and evaluate whether the resulting models can preserve, and in some cases enhance, their reasoning ability.

To ensure fairness, we focus on the scientific domain, where benchmarks are less vulnerable to inflated results previously reported by base AR models. This allows us to better isolate the impact of the modeling paradigm itself. Our findings confirm that SDAR not only maintains the reasoning ability inherited from its AR counterpart but also demonstrates notable gains under test-time scaling strategies such as pass@k and majority voting. These improvements point to an important implication: reinforcement learning—a natural way to optimize reasoning directly—may unlock even greater benefits for SDAR models in the future. Furthermore, by incorporating local bidirectional attention, SDAR relaxes the strictly causal inductive bias of AR models, raising the question of whether this shift harms or benefits reasoning. Our experiments suggest the latter, highlighting SDAR’s promise for complex scientific reasoning and beyond.

\subsection{Experimental Setup}
\paragraph{Training Setup}
Our methodology is designed to build a powerful science-reasoning model, {SDAR-30B-A3B-Sci}, and to rigorously evaluate its capabilities against a comparable autoregressive baseline. Both models originate from the {Qwen3-30B-A3B} checkpoint.

The development of our proposed {SDAR-30B-A3B-Sci} model follows a three-stage pipeline:
\begin{enumerate}
    \item \textbf{Extensive Domain Pre-training (AR Objective):} To instill a deep foundation in scientific knowledge, we first continually pre-train the Qwen3 model using a standard autoregressive objective. This stage leverages a massive 1-trillion-token scientific corpus, which includes a 500B-token general science dataset followed by a 500B-token annealing dataset. This process yields a highly capable, science-aware intermediate AR checkpoint.

    \item \textbf{Paradigm Conversion to SDAR:} Starting from this science-aware AR checkpoint, we perform a paradigm shift. We continue training the model using the blockwise diffusion objective on a curated 50B-token subset sampled from the annealing corpus. This step effectively converts the model's generative mechanism from autoregressive to SDAR, resulting in our base model, {SDAR-30B-A3B-Sci-Base}.

    \item \textbf{Supervised Fine-tuning for Reasoning:} Finally, we fine-tune the {SDAR-30B-A3B-Sci-Base} on a collection of 0.5B high-quality, long CoT instruction datasets to produce the final reasoning model, {SDAR-30B-A3B-Sci}.
\end{enumerate}

To ensure a fair and controlled comparison, we constructed the {AR-30B-A3B-Sci} baseline with a meticulously mirrored training process. It starts from the \textit{exact same} 1T-token science-aware intermediate checkpoint. Then, to isolate the effect of the modeling objective, it is further trained on the \textit{identical} 50B-token data subset using the standard autoregressive objective. This creates the {AR-30B-A3B-Sci-Base}.
Subsequently, it undergoes the same SFT procedure on the identical instruction datasets. This rigorous parallel construction ensures that any observed performance differences are directly attributable to the core modeling paradigm (SDAR vs. AR).

\paragraph{Evaluation Setup}
We conduct a rigorous evaluation of our science-oriented models on a curated suite of frontier benchmarks designed to assess expert-level capabilities in complex reasoning, mathematics, specialized scientific domains, and advanced code generation. The evaluation datasets are organized as follows:

\begin{itemize}
    \item \textbf{Expert Reasoning \& Knowledge:} MMLU-Pro~\citep{wang2024mmlu} and GPQA-diamond~\citep{rein2024gpqa}.
    \item \textbf{Competition-Level Mathematics:} AIME-2024, AIME-2025~\citep{aime25}, and LiveMathBench-hard (LMB-hard)~\citep{liu2024your}.
    \item \textbf{Specialized Scientific Domains:} Chembench~\citep{mirza2024large}, PHYSICS~\citep{feng2025physics}, and ProteinLMBench~\citep{shen2024fine}.
    \item \textbf{Advanced Code Generation:} LiveCodeBench-v5 and LiveCodeBench-v6 (LCB)~\citep{jain2024livecodebench}.
\end{itemize}

For our evaluation protocol, we define distinct decoding strategies for each model architecture to ensure a fair and comprehensive comparison. For our \textbf{SDAR-Sci} models, we use a fixed \texttt{block\_length=4} and static decoding strategy. We report results under two sample strategies: \textbf{(G)} greedy decoding and \textbf{(S)} sampling-based decoding (with \texttt{temperature=1.0}, \texttt{top\_p=1.0}, and \texttt{top\_k=0}). To provide a strong baseline, the \textbf{AR-Sci} model is evaluated using a recommended sampling-based approach with \texttt{temperature=0.6}, \texttt{top\_p=0.95}, and \texttt{top\_k=20}.
For all sampling-based evaluations, including AR-Sci and SDAR-Sci (S), we report the mean performance over multiple runs on key benchmarks: 8 runs for GPQA-diamond, and 32 runs for AIME-2024, AIME-2025, and LiveMathBench-hard.

\paragraph{Test-Time Scaling Setup}
To rigorously assess and augment the reasoning capabilities of our model on complex benchmarks, we employ inference-time strategies. Our standard evaluation protocol across all tasks utilizes Chain-of-Thought (CoT) prompting, compelling the model to externalize its deductive process. Specifically, the model generates its step-by-step reasoning within `<think>` tags before providing the final answer (e.g., `<think>... a detailed chain of thought ...</think>answer`).

For a select subset of reasoning benchmarks, we augment this CoT approach with test-time scaling. We generate multiple diverse reasoning paths ($N=8$ for GPQA-diamond and $N=32$ for AIME-2024, AIME-2025, and LiveMathBench-hard) and report results using two evaluation schemes. The first is \textbf{majority vote}, where the most frequent final answer among the $N$ samples is chosen. The second is the \textbf{pass@k} metric (where $k=N$), which credits a success if at least one of the $k$ generated solutions is correct.
It is crucial to note that, unless explicitly stated otherwise, all results presented in Section~\ref{partiii_results} are derived from the standard CoT protocol alone.

\subsection{Summary of Evaluation Results}\label{partiii_results}
Our comprehensive evaluation shows that the SDAR paradigm aligns well with complex scientific and mathematical reasoning tasks. The clearest evidence comes from the controlled comparison against its autoregressive counterpart, {AR-30B-A3B-Sci}. Both models inherit reasoning ability from the Qwen3 AR base model and re-express it through supervised fine-tuning (SFT)—a less optimal path compared to reinforcement learning (RL), which can instill reasoning ability more directly. Nevertheless, as summarized in Tables~\ref{tab:ar_vs_sdar_nonscience}–\ref{tab:sdar_vs_large_models_category}, SDAR achieves comparable overall performance to AR, with a slight edge in scientific reasoning tasks such as chemistry.

\paragraph{SDAR Paradigm Strengthens Reasoning Capabilities}

On general reasoning and mathematics (Table~\ref{tab:ar_vs_sdar_nonscience}), SDAR performs on par with the AR baseline for MMLU-Pro, AIME-2024, and AIME-2025, but stands out with a strong gain on LiveMathBench-Hard (+5.3\%). In advanced code generation, results are mixed: SDAR achieves a notable improvement on LCB-v6 (+5.1\%) while trailing on LCB-v5, yielding comparable performance on average. These findings suggest that in broad reasoning tasks, SDAR provides stability with selective advantages, signaling untapped potential when paired with more direct optimization methods such as RL.

\begin{table*}[ht]
\centering
\small
\caption{Strict comparison between AR-30B-A3B-Sci and SDAR-30B-A3B-Sci on non-science benchmarks. 
(G = greedy decoding, S = sampling). Bold numbers indicate improvements over the AR baseline, with $\Delta$ shown in a smaller font (green = improvement, red = drop).}
\resizebox{0.95\linewidth}{!}{
\begin{tabular}{l c c c c c c}
\toprule
Model & MMLU-pro & AIME2024 & AIME2025 & LMB-hard & LCB-v5 & LCB-v6 \\
\midrule
AR-30B-A3B-Sci & 
78.3 & 74.9 & 60.7 & 55.4 & 51.5 & 46.3 \\
SDAR-30B-A3B-Sci (G) & 
\valdelta{78.1}{-0.2} & \valdelta{76.7}{+1.8} & \valdelta{\textbf{60.0}}{-0.7} & \valdelta{\textbf{60.7}}{+5.3} & \valdelta{40.7}{-10.8} & \valdelta{42.3}{-4.0} \\
SDAR-30B-A3B-Sci (S) & 
\valdelta{77.9}{-0.4} & \valdelta{\textbf{73.4}}{-1.5} & \valdelta{59.2}{-1.5} & \valdelta{\textbf{58.7}}{+3.3} & \valdelta{49.1}{-2.4} & \valdelta{\textbf{51.4}}{+5.1} \\
\bottomrule
\end{tabular}
}
\label{tab:ar_vs_sdar_nonscience}
\end{table*}

\begin{table*}[ht]
\centering
\small
\caption{Strict comparison between AR-30B-A3B-Sci and SDAR-30B-A3B-Sci on science benchmarks.}
\resizebox{0.8\linewidth}{!}{
\begin{tabular}{l c c c c}
\toprule
\textbf{Model} & \textbf{GPQA-diamond} & \textbf{ChemBench} & \textbf{PHYSICS} & \textbf{ProteinLMBench} \\
\midrule
AR-30B-A3B-Sci & 
61.2 & 60.5 & 39.0 & 59.5 \\
SDAR-30B-A3B-Sci (G) & 
\valdelta{\textbf{66.7}}{+5.5} & \valdelta{\textbf{72.3}}{+11.8} & \valdelta{\textbf{37.9}}{-1.1} & \valdelta{\textbf{59.9}}{+0.4} \\
SDAR-30B-A3B-Sci (S) & 
\valdelta{66.0}{+4.8} & \valdelta{\textbf{72.8}}{+12.3} & \valdelta{\textbf{38.2}}{-0.8} & \valdelta{\textbf{59.6}}{+0.1} \\
\bottomrule
\end{tabular}
}
\label{tab:ar_vs_sdar_science}
\end{table*}

\paragraph{Achieving State-of-the-Art Performance in Scientific Domains}

The science-focused benchmarks (Table~\ref{tab:ar_vs_sdar_science}) reveal SDAR’s most dramatic advantages. It matches AR performance on PHYSICS and ProteinLMBench, but achieves substantial leaps on GPQA-diamond (+5.5\%) and ChemBench (+12.3\%). These results demonstrate that when downstream tasks demand reasoning over structured and associative knowledge—such as chemical formulas—SDAR’s local bidirectional attention mechanism provides a decisive edge (see Appendix ~\ref{appendix:exp}).
Thus, while SDAR proves competitive on general reasoning, it distinctly excels in scientific domains, establishing itself as a powerful alternative to AR for complex, domain-specific applications.

Beyond this AR comparison, {SDAR-30B-A3B-Sci} also proves competitive against much larger state-of-the-art open-source and proprietary models (Table~\ref{tab:sdar_vs_large_models_category}). Despite its relatively modest 30B parameter count, its performance in several scientific domains approaches or even rivals that of models many times its size.

\begin{table*}[ht]
\centering
\small
\caption{Performance comparison of our models (AR-30B-A3B-Sci, SDAR-30B-A3B-Sci), open-source foundation models, and closed-source commercial models.}
\resizebox{0.99\linewidth}{!}{
\begin{tabular}{l l c c c c c c}
\toprule
\textbf{Category} & \textbf{Model} & \textbf{MMLU-pro} & \textbf{AIME2025} & \textbf{GPQA-diamond} & \textbf{ChemBench} & \textbf{PHYSICS} & \textbf{ProteinLMBench} \\
\midrule
\multirow{2}{*}{\textbf{Ours}} 
  & AR-30B-A3B-Sci   & 78.3 & 60.7 & 61.2 & 60.5 & 39.0 & 59.5 \\
  & SDAR-30B-A3B-Sci(S) & \valdelta{77.9}{-0.4} 
                     & \valdelta{59.2}{-1.5} 
                     & \valdelta{\textbf{66.7}}{+5.5} 
                     & \valdelta{\textbf{72.8}}{+12.3} 
                     & \valdelta{38.2}{-0.8} 
                     & \valdelta{\textbf{59.6}}{+0.1}  \\
\midrule
\multirow{4}{*}{\textbf{Open-source}} 
  & InternVL3-78B         & 73.0 & 10.7 & 49.9 & 61.3 & 23.1 & 61.6 \\
  & Qwen2.5-VL-72B        & 72.1 & 10.9 & 49.0 & 61.6 & 15.7 & 61.0 \\
  & DS-R1-0528            & 83.4 & 87.5 & 80.6 & 75.6 & --   & 61.4 \\
  & Qwen3-235B-A22B       & 82.2 & 81.5 & 71.1 & 75.8 & --   & 59.8 \\
\midrule
\multirow{3}{*}{\textbf{Closed-source}} 
  & Gemini-2.5 Pro        & 86.0 & 83.0 & 83.8 & 82.8 & 40.0 & 62.9 \\
  & o3                    & 85.0 & 88.9 & 83.3 & 81.6 & 47.9 & 67.7 \\
  & Grok-4                & 85.9 & 91.7 & 87.5 & 83.3 & 42.8 & 66.2 \\
\bottomrule
\end{tabular}
}
\label{tab:sdar_vs_large_models_category}
\end{table*}

\paragraph{Enhancing Performance with Test-Time Scaling}

\begin{table*}[ht]
\centering
\small
\caption{Strict comparison between AR-30B-A3B-Sci and SDAR-30B-A3B-Sci. 
(G = greedy decoding, S = sampling). Bold numbers indicate best performance in each column among the main models.}
\resizebox{0.8\linewidth}{!}{
\begin{tblr}{
  colspec = {l c c c c}, 
  row{4-6} = {bg=groupcolor}, 
  row{1} = {font=\bfseries},     
}
\toprule
Model & GPQA-diamond & AIME2024 & AIME2025 & LMB-hard \\
\midrule
AR-30B-A3B-Sci & 
61.2 & 74.9 & 60.7 & 55.4 \\
SDAR-30B-A3B-Sci (G) & 
\valdelta{{66.7}}{+5.9} & \valdelta{76.7}{+1.8} & \valdelta{{60.0}}{-0.7} & \valdelta{{60.7}}{+5.3} \\
SDAR-30B-A3B-Sci (S) & 
\valdelta{66.0}{+4.8} & \valdelta{{73.4}}{-1.5} & \valdelta{59.2}{-1.5} & \valdelta{{58.7}}{+3.3} \\
\quad w/ Majority & \valdelta{68.2}{+7.0} & \valdelta{86.7}{+11.8} & \valdelta{80.0}{+19.3} & \valdelta{71.1}{+15.7}\\
\quad w/ pass@k & \valdelta{84.3}{+23.1} & \valdelta{93.3}{+18.4} & \valdelta{86.7}{+26.0} & \valdelta{87.5}{+32.1}\\
\bottomrule
\end{tblr}
}
\label{tab:ar_vs_sdar_delta_transposed_subset}
\end{table*}

The application of test-time scaling techniques further amplifies SDAR's inherent reasoning advantages, revealing its exceptional potential when combined with advanced inference strategies. As demonstrated in Table~\ref{tab:ar_vs_sdar_delta_transposed_subset}, our SDAR model exhibits remarkable performance gains when augmented with ensemble methods. The majority voting approach yields substantial improvements across all benchmarks, with particularly striking gains on mathematical reasoning tasks: \textbf{+11.8\%} on AIME-2024, \textbf{+19.3\%} on AIME-2025, and \textbf{+15.7\%} on LiveMathBench-hard compared to the AR baseline. Even more impressive are the results achieved with the pass@k methodology, which pushes performance to unprecedented levels—reaching \textbf{84.3\%} on GPQA-diamond (\textbf{+23.1\%} over baseline), \textbf{93.3\%} on AIME-2024 (\textbf{+18.4\%}), \textbf{86.7\%} on AIME-2025 (\textbf{+26.0\%}), and \textbf{87.5\%} on LiveMathBench-hard (\textbf{+32.1\%}). These improvements highlight an important observation: SDAR’s parallel generation paradigm tends to produce more diverse and complementary reasoning trajectories when sampling multiple candidates, suggesting stronger synergy with test-time scaling strategies such as majority voting or pass@k evaluation. This interplay between SDAR’s architectural design and test-time scaling methods points to a promising new direction for enhancing machine reasoning efficiency—indicating that the combined effect of these techniques can extend beyond what either alone achieves in isolation.

\section{Related Work}
\label{sec:related_work}

Our work, {SDAR}, resides at the intersection of autoregressive and diffusion-based language modeling, aiming to synthesize their respective strengths while mitigating their inherent weaknesses. We structure our review by first examining these two dominant paradigms, then analyzing prior attempts to unify them, and finally contrasting our approach with orthogonal methods for inference acceleration.

\subsection{The Autoregressive Paradigm: Dominance and Inherent Constraints}

Autoregressive (AR) models, which factorize the joint probability of a sequence into a left-to-right product of conditionals, represent the de facto standard in large-scale language modeling~\citep{radford2019language,achiam2023gpt,grattafiori2024llama,yang2025qwen3}.
The success of this paradigm is rooted in its strict causal inductive bias, which aligns naturally with the sequential structure of human language~\citep{bengio2003neural,vaswani2017attention,radford2018improving}.
This alignment, combined with the stable and efficient cross-entropy training objective, has enabled unprecedented scaling and performance on a wide array of general-purpose tasks~\citep{kaplan2020scaling,brown2020language}.

However, the very causal bias that underpins the AR model's success becomes its \textbf{Achilles' heel} for tasks demanding non-local or holistic reasoning. This fundamental mismatch has been shown to impede performance on problems where solutions depend on global constraints, such as mathematical puzzles (e.g., Sudoku, 24-point game), Boolean satisfiability, and long-term planning~\citep{hu2023amortizing, bachmann2024pitfalls, ye2024beyond}. 
The model's rigid, token-by-token generation process struggles to reason about future dependencies or revise earlier decisions in light of global information. 
For instance, recent studies demonstrate that masked generative diffusion models can dramatically outperform strong AR baselines on combinatorial problems like Sudoku (e.g., 91.5\% vs. 45.8\% accuracy~\citep{ye2024beyond}), underscoring a clear paradigm-level advantage. 

This limitation also manifests in scientific domains. Modeling chemical formulas (e.g., SMILES), for instance, requires bidirectional context. Similarly, modeling protein sequences to resolve structural motifs and functional domains is inherently at odds with the unidirectional autoregressive (AR) process, a foundational challenge for generative models like the ProGen series~\citep{madani2023large, nijkamp2023progen2, bhatnagar2025scaling}. This has motivated a shift towards architectures that can leverage global, bidirectional context. Examples include powerful representation learners like the ESM series~\citep{meier2021language,hsu2022learning,lin2022language} and non-autoregressive generators such as discrete diffusion models dplm2~\citep{wang2024dplm}. As argued by~\citet{liu2025sentences}, the standard sequential paradigm is fundamentally suboptimal for biological sequences due to their critical long-range dependencies. Our own experiments affirm this hypothesis: by locally relaxing this causal constraint, {SDAR} achieves a 72.8\% accuracy on the challenging Chembench benchmark~\citep{mirza2024large}, significantly outperforming its 60.5\% pure AR counterpart.

\subsection{Diffusion Models: Holistic Modeling at a Prohibitive Cost}

As an alternative, discrete diffusion models circumvent the strict causal constraints of AR models by treating sequence generation as a holistic denoising process, which learns to reverse a gradual corruption of the data~\citep{strudel2022self,han2022ssd,gulrajani2023likelihood,shi2024simplified,lou2024discrete}.
This architectural freedom offers two profound advantages: (1) a natural capacity for parallel decoding, directly addressing the latency bottleneck of AR inference, and (2) a flexible, bidirectional inductive bias that is better suited for the aforementioned non-local reasoning tasks.

Despite this promise, the practical application of diffusion language models has been severely hampered by their prohibitive training cost~\citep{nie2024scaling,gulrajani2023likelihood}. 
This inefficiency stems from two primary sources. First, the training objectives, such as the Evidence Lower Bound (ELBO), are often more difficult to optimize and converge slower than the standard cross-entropy loss used in AR models~\citep{nie2024scaling, arriola2025block}.
Second, the learning task of reconstructing a fully structured sequence from varying levels of noise is information-theoretically more challenging than next-token prediction. 
Empirical studies quantify this gap, showing that masked discrete diffusion models can require up to $16\times$ more FLOPs to match the validation NLL of an AR equivalent~\citep{nie2024scaling}, a disparity that widens to as much as $64\times$ for continuous diffusion models~\citep{gulrajani2023likelihood}.

\subsection{Hybrid and Conversion Models: Bridging the Gap}

Recognizing the complementary nature of AR and diffusion models, several lines of research have explored hybrid architectures.

\paragraph{Block-wise Hybrid Models.}
A prominent approach fuses the paradigms by employing an autoregressive structure at a global, inter-block level while using a parallel, diffusion-based mechanism for intra-block generation~\citep{arriola2025block, han2022ssd, fathi2025unifying}. This strategy elegantly preserves macroscopic causal flow, enabling variable-length generation and KV-caching. However, these models have traditionally adopted a monolithic training regime, where the AR and diffusion components are trained jointly from scratch.
This subjects them to the full training inefficiency of the diffusion objective, compounded by the complexity of a hybrid loss function, presenting a significant barrier to scaling.

\paragraph{AR-to-Diffusion Conversion Models.}
Another strategy attempts to leverage the efficiency of AR pre-training by first training a standard LLM and then adapting it to a non-autoregressive or diffusion-based objective. Works like DiffuLLaMA~\citep{gongscaling} and Dream 7B~\citep{ye2025dream} follow this path. While conceptually appealing, this conversion has proven to be computationally expensive (e.g., requiring 580B tokens for adaptation in Dream 7B) and can result in notable performance degradation compared to the original AR model.
In stark contrast, {SDAR}'s \textbf{decoupled training and inference paradigm} and block-level adaptation requires a minimal fine-tuning budget (e.g., 50B tokens) to unlock parallel decoding, while fully preserving, and on specialized tasks enhancing, the performance of the foundational AR model.

\section{Conclusion}
This work presents SDAR, a Synergistic Diffusion–AutoRegression paradigm that redefines the design space of large language models by decoupling training efficiency from inference parallelism. Through a unified framework integrating autoregressive pretraining with blockwise diffusion inference, SDAR provides a scalable and practical pathway beyond the limitations of purely autoregressive modeling.

Our analyses confirm that autoregressive training remains the most compute-efficient formulation for large language model. Building upon this foundation, SDAR performs a lightweight adaptation that converts a well-trained autoregressive model into a blockwise diffusion model, thereby preserving AR-level performance while introducing parallel intra-block decoding. This adaptation retains critical AR functionalities—such as variable-length generation and KV caching—while alleviating the sequential bottleneck and restrictive causal inductive bias of token-by-token decoding.

Through extensive controlled experiments, we demonstrate that SDAR models maintain the performance of their autoregressive counterparts under identical computational budgets, validating the feasibility and robustness of the adaptation process. Scaling experiments across both dense and MoE architectures further show that the SDAR paradigm scales without compromise: as model capacity increases, SDAR models preserve accuracy and exhibit increasingly higher parallel decoding efficiency. Larger models tolerate greater block sizes and more aggressive decoding thresholds, revealing a virtuous cycle of scaling in which enhanced model capability directly translates into improved computational throughput.

Beyond efficiency, SDAR demonstrates strong reasoning and domain-specific capabilities, particularly in scientific and mathematical contexts that benefit from localized bidirectional context. Our findings show that SDAR not only preserves the reasoning strength of its AR foundation but also achieves substantial gains under test-time scaling strategies such as majority voting and pass@k, highlighting its compatibility with reinforcement learning and other post-training optimization techniques.

In summary, SDAR is not merely a hybrid architecture, but a new language modeling paradigm that unifies the strengths of autoregression and diffusion while mitigating their respective limitations. It retains the optimization efficiency and controllability of AR models, inherits the holistic representation capacity of diffusion, and achieves scalable, parallelizable generation without loss of quality. We believe SDAR establishes a principled and extensible foundation for the next generation of language models—one that moves beyond the autoregressive frontier toward a broader spectrum of efficient and synergistic modeling paradigms.

\section{Acknowledgements}
We thank Shen Nie for insightful discussions on the training and evaluation details of LLaDA, and for suggesting that the training transition could proceed without logits shift or attention mask annealing. We thank Yu Zhang for valuable guidance in optimizing the pretraining pipeline and training framework, as well as for providing infrastructure support in the early stages of the project.

\clearpage
\newpage
\bibliographystyle{plainnat}
\setcitestyle{numbers}
\bibliography{paper}

\clearpage
\beginappendix
\section{Infrastructure}
\subsection{Training Infrastructure}
The training of the SDAR model requires specialized attention masks and training objectives that differ from standard autoregressive language models. As a result, widely adopted kernels such as FlashAttention for attention computation and fused cross-entropy kernels from projects like Liger-Kernel cannot be directly applied. To address this, we adopt FlexAttention\footnote{\url{https://pytorch.org/blog/flexattention/}}, which achieves significant speedups compared to the scaled dot-product attention provided in PyTorch.

For the loss computation, we implement a custom Triton kernel that performs a fused operation to calculate the training loss. This design avoids fully materializing the logits tensor—a highly memory-intensive step—thereby reducing GPU memory usage and improving overall efficiency.

While the full-scale, industry-grade training framework we use is not entirely open source, we plan to release a research-oriented open-source implementation based on existing widely used frameworks. In terms of efficiency, the raw throughput, measured in tokens processed per GPU per second, is comparable to conventional autoregressive model training. However, due to SDAR’s training paradigm,which requires processing both clean and noised versions of each sample, the effective training speed is approximately halved relative to standard autoregressive models.

\subsection{Inference Infrastructure}
Unlike many diffusion-based models that rely on approximate or specialized cache mechanisms (e.g., dLLM-cache), the SDAR model maintains a mathematically identical key-value (KV) cache to that of standard autoregressive models. This compatibility allows us to leverage established cache management techniques, including PagedAttention as implemented in vLLM.

We design a custom attention kernel to handle the specialized attention mask required during the prefill stage. For the decoding stage, we integrate PagedAttention~\citep{kwon2023efficient}, ensuring inference efficiency that is competitive with state-of-the-art autoregressive inference engines. Our implementation extends autoregressive infrastructure with block-oriented optimizations, employing block-aligned memory allocation that provisions $(n/n+1)$ blocks to support $n$ forward passes. The inference pipeline operates through iterative resource allocation, batched block-level inference, and state updates. Key optimizations include unified treatment of denoising and cache-filling operations during forward passes, fully batched unmasking operations to avoid device synchronization, and deferred stop condition evaluation to handle mid-block termination sequences. Beyond efficiency, the JetEngine\footnote{\url{https://github.com/Labman42/JetEngine}} is intentionally developed to be lightweight and modular, making it well-suited for future research directions such as integrating novel sampling strategies and reinforcement learning–based algorithms. 

We thank the lmdeploy team for integrating SDAR into their industrial-grade inference engine,\footnote{\url{https://github.com/InternLM/lmdeploy}} enabling prefix-based KV-cache sharing across requests and thereby lowering computational and memory overheads for interactive generation.

\section{Example Outputs}
\label{appendix:exp}
\begin{tcolorbox}[title=SDAR-30B-A3B-Sci Example Output for chemistry problem, breakable]
\textbf{User:}


There is a single choice question about chemistry. Answer the question by replying A, B, C or D.\\
Question: Generate a molecule based on this description.\\
The molecule is an oligosaccharide.\\
A. I believe the answer is c1([nH]c2c(c3c1c(c1c(c4c(cc(C)cc4)nc1c3OC)OC)O)cccc2)=O\\
B. I believe the answer is CC(=O)OC1C23C(C(OC(C)=O)C(OC(=O)c4ccccc4)CC2C)\\(COC(=O)c2ccccc2)C(OC(c2ccccc2)=O)C(OC(C)=O)C1C(O3)(C)C\\
C. As for my answer, it is C(CCCCC(OCC(OC(=O)CCC=CCC=CCC=CCC=CCC=CCC=CCC)\\COP([O-])([O-])=O)=O)CCC=CCCCCCCCC\\
D. In my opinion, the answer would be \\C1(CC(OC2C(C(OC(C2)OC2CC(OC34C(c5c(O)c6c(O)c(C)c(O)cc6cc5CC4C\\
(C(=O)C(C(C)=O)=C3O)OC)=O)OC(C2O)C)C)O)OC(C)C1O)(C)O\\
Answer: \\

\textbf{Assistant:}


<think>\\
Okay, so I need to figure out which of these four options is an oligosaccharide. Let me start by recalling what an oligosaccharide is. Oligosaccharides are carbohydrates composed of three to ten monosaccharide units linked together by glycosidic bonds. Monosaccharides are simple sugars like glucose, fructose, galose, etc. So the molecule should be a short chain of these sugar units.\\

Looking at the options given: A, B, C, D. Each seems like a complex string of molecular symbols. Let's parse each one.\\

Option A is a complicated structure with multiple rings. The formula starts with c1([h2]c2...). It looks like there's a benzene ring involved, maybe some heterocyclic rings. Hmm, there's nitrogen here. Wait, adenine or something? No, in the context of a sugar, maybe not. Wait, the description says it's an oligosaccharide. If this molecule has a nitrogen-containing sugar, like N-acetylglucosamine, that could be part of a glycosaminoglycan, but is this option an oligosaccharide? Not sure yet. Let me check the other options first.\\

Option B has a lot of acetyl groups (O2CH3), benzoyl groups (OC(=O)C6H5). It seems like a complex ring structure with many substituents. There's a lot of rings and a phosphate group (O-P([-)(O)=O) in the structure. Wait, phosphate might indicate part of a nucleotide or something else, not an oligosaccharide. Maybe, but not sure. The presence of a phosphate would make it a phosphoryated sugar derivative, which could be part of a larger molecule but not alone oligosaccharide.\\

Option C is a very long chain with multiple double bonds, probably a fatty acid or a glycerophospholipid. Wait, the formula starts with C(C...C(OCC(...COP([O-])([O-]))=O). There's a phosphate group again, and a long hydrocarbon tail. This looks more like a phospholipid, which is part of cell membranes. Definitely not an oligosaccharide.\\

Option D: a molecule with multiple sugar rings. Let's parse it. The structure starts with C1(C(C)OC2...). There's a lot of oxygen rings, possibly pyranose rings (six-membered sugar rings). There are a lot of oxygen atoms connected to carbon rings, which is typical of sugars. The molecule has multiple sugar units connected by oxygen bridges. For example, the presence of -OC(C2O)C... suggests glycosidic linkages. Also, there are several O-C attached to other sugar rings. Let me count: oligosaccharides have multiple monosaccharide units. If this molecule has two such rings connected via a glycosidic bond, then it would be an oligosaccharide. The structure in option D seems complex and has multiple sugar rings connected by oxygen atoms. The presence of multiple oxygen bridges and six-membered rings (like pyranose) and five-membered rings (furanose) would be consistent with an oligosaccharide.\\

Wait, but let's check each option again. Option D's structure starts with a central carbon (C1) with several substituents, including a ring (OC2). The second ring (OC2) is also connected to other rings (OC34C...). The "OC34" suggests a bridge, maybe a disaccharide or more. The structure seems to have sugar units connected via glycosidic bonds. For instance, the presence of "OC(C2O)C" might indicate a connection between two sugar molecules. Also, the multiple hydrox groups (-OH) are common in sugars. The structure in D seems to have several monosaccharide units linked together.\\

Option A's structure looks like a complex polycyclic aromatic system with nitrogen. Maybe a purinergic alkaloid, but not a sugar. Option B's structure has a lot of acetyl and benzoyl groups, which are typical in sugar derivatives (like N-acetylneurosamine?), but the structure seems to be a complex ring with more than one sugar. Wait, option B might be a glycoside, but how many sugars? Let me think: the structure in B has a hexose ring with multiple ester groups (O2(=O) and benzoyl groups. But is that a single sugar or multiple oligosaccharide? If one sugar with substituents, then it's a monosaccharide derivative, not an oligosaccharide. Option D has more evidence of multiple linked sugar units.\\

Option D is a bit hard to parse, but let me look again. The formula starts with "C1(CC(OC2C(C(O(C2O)C...))..." which seems like a first sugar ring (C1), connected via an oxygen to another ring (OC2). That's a glycosidic linkage. If there are two rings connected like this, then it's a disaccharide. If there are more connections, more units. The presence of multiple such connections would indicate oligosaccharide. Also, in option D, there's "OC(C2O)C" which might be a linkage between two sugars. Also, there are multiple O's in the structure. So this seems more likely to be an oligosaccharide.\\

Options A and B don't seem to have multiple rings. Option C is a lipid. So the answer is D.\\
</think>\\

The correct answer is D because the molecule described in this option contains multiple sugar units (pyranose and furanose rings) connected by glycosidic bonds, which is the defining feature of an oligosaccharide. The presence of oxygen bridge linkages between these monosaccharide units confirms its classification as an oligosaccharide. Other options either represent single sugar derivatives (B), acyclic or aromatic structures (C, A), or are not composed of monosaccharide units entirely.\\

Answer: D\\

\end{tcolorbox}

\begin{tcolorbox}[title=SDAR-30B-A3B-Sci Example Output for math problem, breakable]
\textbf{User:}


The 9 members of a baseball team went to an ice cream parlor after their game. Each player had a singlescoop cone of chocolate, vanilla, or strawberry ice cream. At least one player chose each flavor, and the number of players who chose chocolate was greater than the number of players who chose vanilla, which was greater than the number of players who chose strawberry. Let $N$ be the number of different assignments of flavors to players that meet these conditions. Find the remainder when $N$ is divided by 1000.\\
Remember to put your final answer within \\boxed{}.\\

\textbf{Assistant:}


<think>\\
Okay, so there's this problem about a baseball team with 9 members. They all went to an ice cream parlor and each had one scoop of chocolate, vanilla, or strawberry. The conditions are that each flavor was chosen by at least one person, and the number of chocolate lovers is greater than vanilla, which is greater than strawberry. We need to find the number of possible assignments (N) that meet these conditions and then find N mod 1000.

Hmm, let's break this down. First, we have 9 players, each choosing one of three flavors. The constraints are:

1. Each flavor is chosen by at least one person.\\
2. Chocolate > Vanilla > Strawberry.

We need to count all possible triples (C, V, S) where C + V + S = 9, C, V, S >= 1, and C > V > S >= 1. Then, for each valid triple, calculate the number of ways to assign the flavors to the players, which is a multinomial coefficient.

Wait, actually, first, we have to find all possible triples (C, V, S) of positive integers such that C > V > S and C + V + S = 9. Then, for each such triple, compute the number of assignments, which is 9! / (C! V! S!), and sum all those up. Then take N mod 1000.

Okay, so step 1: Find all triples (C, V, S) with C > V > S >= 1 and C + V + S = 9.

First, let's find all integer partitions of 9 into three distinct parts, each part at least 1, in descending order.

Let me think. Let's list all possible combinations. Since C > V > S >= 1, and S must be at least 1, then S >= 1, V >= S + 1, C >= V + 1. Also, C + V + S = 9.

Let’s denote S >= 1, V >= S + 1, C >= V + 1. So substituting:

C + V + S >= (V + 1) + V + S = 2V + S + 1. But since C + V + S = 9, then 2V + S + 1 <= 9.

But maybe a better way is to iterate over possible values of S.

Since S must be at least 1, and V and C must be smaller and larger than S respectively. Let's consider S starting at 1.

Case S = 1:

Then V must be at least S + 1 = 2. Then C must be at least V + 1 = V + 1. And C + V = 8.

So C + V = 8, with C > V >= 2.

Let’s list possible V:

If V = 2, C = 6. Since 6 > 2 > 1, this works.

V = 3, C = 5. 5 > 3 > 1.

V = 4, C = 4. But then C = V, which doesn't satisfy C > V. So stop here.

So for S = 1, two triples: (6,2,1) and (5,3,1).

Case S = 2:

Then V >= 3, C >= V + 1. C + V = 7.

Possible V:

V = 3: C = 4. 4 > 3 > 2. Okay.

V = 4: C = 3. But C < V here. Doesn't work. So only (4,3,2).

Case S = 3:

Then V >= 4, C >= V + 1. C + V = 6. But V >= 4, so V = 4, C = 2. But C < V. Not possible. So no solutions here.

Wait, S = 3, must have V > S, so V >= 4, and C >= V + 1 >= 5. Then C + V >= 5 + 4 = 9. But C + V = 9 - S = 6. So 6 >= 9? Not possible. Therefore, S cannot be 3 or more.

Therefore, all possible triples are:

(6,2,1), (5,3,1), (4,3,2).

Wait, hold on. Let's check again.

Wait, S = 2:

C + V = 9 - 3 = 6? Wait, wait, S = 2, so C + V = 9 - S = 7.

Wait, S = 2: C + V = 9 - 2 = 7. Yes, that's right. So V = 3, C = 4. 4 + 3 = 7, correct.

Wait, S = 1: C + V = 8, so V = 2, C = 6; V = 3, C = 5.

S = 2: C + V = 7, V = 3, C = 4.

S = 3: C + V = 6. But V must be 4, so C = 2, which is invalid.

So only three triples:

(6,2,1), (5,3,1), (4,3,2). Wait, wait, wait. Wait, (4,3,2): C=4 > V=3 > S=2. Correct.

Wait, so 6 + 2 + 1 = 9; 5 + 3 + 1 = 9; 4 + 3 + 2 = 9. Correct.

So only three triples.

Wait, but wait, is there another case when S=1, V=4? Wait, if V=4, then C=4, but C must be greater than V, so no. So yes, only three triples.

Wait, but hold on, let me check again. Let's list all possible triples with S >= 1, C > V > S.

Alternatively, since all flavors are distinct, and with order C > V > S, so S must be at least 1, V at least S+1, C at least V +1.

But another way: let's think of the numbers as three distinct positive integers in descending order adding to 9.

So partitions of all permutations of 9 into 3 parts, each part >=1, decreasing.

Wait, how would that go?

Let me list all partitions of 9 into 3 distinct parts.

Wait, partitions with distinct parts:

Start with the largest possible part.

First partition: 6,2,1. 6+2+1=9.

Next: 5,3,1. 5+3+1=9.

Next: 4,3,2. 4+3,2=9.

Is there another?

Wait, 4,4,1? Not distinct. 5,2,2? Not distinct. 3,3,3? Not distinct. 7,1,1? Not. So yes, only three triples.

Therefore, only three valid triples.

Therefore, the total number of assignments is the sum over these three triples of the multinomial coefficients.

So for each triple (C, V, S), the number of assignments is 9! / (C! V! S!)!.

Therefore:

First triple: (6,2,1). Assign: 9! / (6! 2! 1!) = (362880)/(720 * 2 * 1) = 362880 / 1440 = 252.

Wait, 6! = 720, 2! = 2, 1! = 1. So denominator 720*2=1440. 9! = 362880. 362880 / 1440: Let's compute.

Divide numerator and denominator by 100: 362880 / 1440 = 3628. / 14.4? Wait, maybe better:

Divide numerator and denominator by 10: 36288 / 144.

Then divide numerator and denominator by 12: 3024 / 12.

Wait, 144 * 200 = 28800. 36288 - 28800 = 7488. 144 * 50 = 7200. 7488 - 7200 = 288. 144*2=288. So total 200 + 50 + 2 = 252. Correct.

Second triple: (5,3,1). 9! / (5! 3! 1!) = 362880 / (120 * 6 * 1) = 362880 / 720 = 504.

Wait, 5! = 120, 3! = 6, so denominator 120*6=720. 362880 / 720 = 504. Correct.

Third triple: (4,3,2). 9! / (4! 3! 2!) = 362880 / (24 * 6 * 2) = 362880 / 288.

Let me compute 362880 / 288.

Divide numerator and denominator by 10: 36288 / 28.8.

Wait, better divide numerator and denominator by 24: 362880 /24 = 15120; 288 /24 = 12. So 15120 / 12 = 1260.

Wait, 24*6 = 144. Wait, no:

Wait, 4! = 24, 3! = 6, 2! = 2. So denominator: 24*6*2 = 288.

362880 / 288: Let's do 362880 divided by 288.

Divide numerator and denominator by 10: 36288 / 28.8. Not helpful.

Alternatively, 288 * 1000 = 288,000. 362,880 - 288,000 = 74,880.

288 * 200 = 57,600. 74,880 - 57,600 = 17,280.

288 * 60 = 17,280. So total: 1000 + 200 + 60 = 1260. Correct.

Therefore, the three multinomial coefficients: 252, 504, and 1260. Summing them: 252 + 504 = 756; 756 + 1260 = 2016.

Wait, so the total number N = 2016. Then the problem asks for N mod 1000, so 2016 divided by 1000 is 2 remainder 16. So 2016 mod 1000 = 16.

Wait, but hold on, that seems low. Wait, 2016 mod 1000 is 16. Hmm.

But wait, let me check again. Did I miss all possible triples?

Wait, but according to my earlier analysis, there are only three triples. Let me just make sure that's correct.

If S=1, then V can be 2 or 3.

If V=2, C=6; V=3, C=5.

If S=2, then V=3, C=4.

If S=3, impossible.

But wait, could there be triples where S=1, V=4, but C=4? But then C would equal V, which is not allowed. Similarly, V=4, S=1: V < C.

Wait, another thought: what if S=1, V=1? But no, V must be greater than S, so V >= 2.

Wait, okay, seems only three triples.

Wait, but wait, another way to think: think of the number of integer solutions to C + V + S = 9 with C > V > S >= 1.

This is equivalent to the number of integer solutions with C > V > S >= 1.

Let me think of substitution variables.

Let’s set:

Let a = S >= 1

Let b = V - S >=1 (since V > S, so V - S >=1)

Let c = C - V >=1 (since C > V, so C - V >=1)

Then C = V + c, V = S + b, so C = S + b + c.

So C + V + S = (S + b + c) + (S + b) + S = 3S + 2b + c = 9.

With a >= 1, b >=1, c >=1, and S = a.

Wait, so substituting variables:

S = a >=1

b = S + b' something? Wait, maybe my initial substitution approach is complicating.

Alternatively, since C > V > S >=1, we can let:

Let’s define S = S

1 <= S < V < C

And C + V + S = 9.

Let’s shift variables as:

Let S' = S - 1 >=0

V' = V - 2 >=0 (since V >= S + 1 >= 2)

C' = C - 3 >=0 (since C >= V +1 >= 3)

Wait, but this might not work because V and C depend on S.

Alternatively, use substituting variables for the differences:

Let’s define:

Let’s let x = S >=1

Let y = V - S >=1 (since V > S)

Let z = C - V >=1 (since C > V)

Therefore:

C = S + y + z

V = S + y

So C + V + S = (S + y + z) + (S + y) + S = 3S + 2y + z = 9

With x >=1, y >=1, z >=1.

But x = S >=1, so we can write substituting variables:

Let S’ = S - 1 >=0

y’ = y -1 >=0

z’ = z -1 >=0

Then:

3(S’ +1) + 2(y’ +1) + (z’ +1) = 9

=> 3S’ + 3 + 2y’ + 2 + z’ +1 = 9

=> 3S’ + 2y’ + z’ + 6 = 9

=> 3S’ + 2y’ + z’ = 3

Now we need to find all non-negative integers S’, y’, z’ such that 3S’ + 2y’ + z’ = 3.

This equation can be solved by enumerating possible values of S’.

Case S’ =0:

Then 2y’ + z’ =3

Possible y’ =0: z’=3

y’=1: z’=1

y’=2: 2*2=4 >3, no.

So two solutions.

Case S’ =1:

3*1 + 2y’ + z’ =3 => 2y’ + z’=0 => y’=0, z’=0.

One solution.

S’ >=2: 3S’ >=6 >3, no.

So total three solutions:

1. S’=0, y’=0, z’=3 => S=1, y=1, z=4. Then:

S = S + y + z = 1 + 1 +4 =6? Wait, wait:

Wait, original variables:

Wait, S = S’ +1, right?

Wait, S’ = S -1, so S = S’ +1.

Similarly, y = y’ +1, z = z’ +1.

So:

Case 1: S’=0, y’=0, z’=3:

S =1, y=1, z=4

Then C = S + y + z =1 +1 +4 =6

V = S + y =1 +1=2

So (6,2,1). Correct.

Case 2: S’=0, y’=1, z’=1:

S=1, y=2, z=2

C =1 +2 +2=5. V =1 +2=3. So (5,3,1). Correct.

Case3: S’=0, y’=0, z’=3?

Wait, wait: Wait, for S’=0:

First solution was y’=0, z’=3.

Second solution y’=1, z’=1.

Third, y’=2 gives no solution.

Wait, for S’=1:

y’=0, z’=0:

S=1 +1=2, y=0 +1=1, z=0 +1=1.

Wait, S=2, V = S + y =2 +1=3, C = S + y + z =2 +1 +1=4. So (4,3,2). Correct.

Therefore, exactly three triples. So my analysis of triples was correct.

Thus, the total number of assignments is indeed 252 + 504 + 1260 = 2016.

So 2016 mod 1000 is 16.

Wait, but wait, hold on. Wait, is this correct? Because the problem says \"the number of different assignments of flavors to players\".

But in multinomial coefficients, each permutation is counted differently. Wait, but in our case, we are assigning different counts to different flavors. So the multinomial coefficient counts the number of ways to divide the 9 players into groups of size C, V, S. But since the flavors are distinguishable (chocolate, vanilla, strawberry), each group's assignment to a flavor is fixed.

Wait, actually, in the multinomial coefficient, 9! / (C! V! S!) counts the number of ways to assign each player a flavor of type C, V, S. Wait, actually, no. Wait, multiple players can have the same flavor.

Wait, actually, the multinomial coefficient formula counts the number of ways to partition the 9 players into a group of size C (chocolate), a group of size V (vanilla), and S (strawberry). So since the flavors are fixed, that's exactly the number of assignments. So yes, that calculation is correct.

But wait, but in the problem statement, the constraints are on the counts: C > V > S. So we already considered that when finding the triples, so the multinomial coefficients are correct.

Therefore, total number N = 2016. 2000 / 1000 = 2, remainder 16. So answer is 16.

Wait, but wait, I feel a bit uncertain because 2016 seems a bit low, but given that the multinomial coefficients are not that big. Let me check:

Wait, 252 + 504 = 756, plus 1260 = 2016. Hmm, okay.

Wait, but wait, the total number of assignments without any constraints (except at least one of each flavor) would be $3^9 - 3*2^9 + 3*1^9$ = using inclusion-exclusion. Let's compute that.

$3^9 = 19683$

$2^9 = 512$, so $3*512 = 1536$

$1^9 =1$, so $3*1=3$

Thus total is 19683 - 1536 + 3 = 19683 - 1533 = 18150.

But our N is 2016, which is less than 18150. So that seems possible, since we have a lot more constraints.

Alternatively, if we consider the number of solutions to C + V + S =9 with C, V, S >=1, that's C(8,2)=28. Then, with C > V > S, we have 3 triples, which is 3/28. Then 18150 total assignments with at least one of each, times 3/28 is roughly (18150 / 28) *3 $\approx$ 648 *3 = 1944, which is close to 2016. So that seems plausible.

But wait, actually, the assignments with all three flavors present and C > V > S would be equal to the number of ordered triples times the multinomial coefficients, but actually, no. Wait, no, each ordered triple (C, V, S) with C > V > S corresponds to exactly one multinomial coefficient. But in reality, in the multiset, the assignments are counted based on the number of people per each flavor. So if you have different counts of C, V, S, then each permutation of the counts would correspond to different assignments. Wait, but in our problem, the flavors are fixed. That is, the assignment must have a certain number of chocolate, vanilla, strawberry, with the counts C > V > S. Therefore, the multinomial coefficient correctly counts the number of assignments for each triple. So yes, 2016 seems okay.

Wait, but wait, in the problem statement, it says \"the number of different assignments of flavors to players\". So each assignment is a function from the 9 players to the 3 flavors, with the constraints that each flavor is used at least once, and the number of chocolate > vanilla > strawberry.

Therefore, our approach is correct. So N is 2016, so mod 1000 is 16. Hmm.

Wait, but just to make sure, let's see if there's another way this problem could be interpreted.

Wait, for example, if we thought that assignments are different if the counts are different, but considering the permutations. Wait, but no, in combinatorics, assignments are usually different if a player has a different flavor. So each assignment is a mapping of each player to one of the three flavors, so the entire distribution matters. So the multinomial coefficient approach is correct.

Wait, but let me think again. Suppose we have three people: Player A, B, C. If all three choose different flavors, then the number of assignments is 3! = 6. Similarly, in our case, the multinomial coefficient counts all possible distributions where the counts are fixed. So in our problem, for each triple (C, V, S), the number of assignments is indeed 9! / (C! V! S!). So the calculation seems correct.

Wait, so the answer should be 2016 mod 1000 = 16. Hmm, but I have a doubt because sb qbq maybe when we consider that the counts must be C > V > S, but the problem says \"assignments\", so maybe if two assignments have the same counts but different flavor assignments? Wait, no, because the flavors are fixed. Wait, actually, no. Wait, actually, chocolate > vanilla > strawberry. Wait, wait, hold on.

Wait a second. Wait, in the problem statement, the number of players who chose chocolate was greater than vanilla, which was greater than strawberry. So the counts must satisfy C > V > S, but the flavors are assigned to these counts. Wait, but in our analysis, we assumed that the counts C, V, S correspond to the flavors chocolate, vanilla, strawberry. Wait, but does the problem specify that the flavors are fixed? That is, if a player has a chocolate cone, that's fixed. But in our counts, we have considered the number of assignments to the counts C, V, S where C > vanilla, which is greater than strawberry. Wait, no, actually, in our problem, the counts C, V, S correspond to the actual counts of chocolate, vanilla, strawberry. Wait, actually:

Wait, no. Wait, hold on. Wait, confusion here.

Wait, let's re-express:

Let me clarify. Let me denote:

Let C = number of players who chose chocolate

V = number of players who chose vanilla

S = number of players who chose strawberry

Given that C > V > S >=1, and C + V + S =9.

Therefore, when we compute the multinomial coefficient, we are calculating the number of ways to assign C players to chocolate, V to vanilla, and S to strawberry. Since each assignment is a mapping from players to flavors with exactly those counts. Therefore, our calculations are correct.

Therefore, the total N = 252 + 504 + 1260 = 2016, so answer 16.

Wait, but in the problem statement, it says \"the number of different assignments of flavors to players\". So \"assignments\" here are the different functions, where each player is assigned one flavor. Then, if two assignments have the same counts but different shuffles, they are different. So yes, the multinomial coefficient counts all such assignments.

Therefore, I think my answer is correct.

Wait, but just to confirm, let's check an example. Suppose we have C=4, V=3, S=2.

Then the number of assignments is 9! / (4! 3! 2!) = 1260. Which is correct, because you can choose 4 people for chocolate, then 3 out of remaining 5 for vanilla, and the last 2 are strawberry. Which is C(9,4)*C(5,3)*C(2,2) = 126 * 10 *1= 1260. Correct.

Similarly, for (6,2,1), C(9,6)*C(3,2)*C(1,1) = 84 * 3 *1 = 252. Correct.

Same with (5,3,1): C(9,5)*C(4,3)*C(1,1) = 126 *4 *1 = 504. Correct.

So all counts are correct.

Therefore, the total N = 2016, 2016 mod 1000 is 16.

Wait, but wait, I have a doubt because in some problems, similar ones have different answers. Wait, let me check with another approach.

Alternatively, we can use generating functions.

But not sure. Alternatively, think of the number of ordered triples with C > V > S >=1, and compute the sum of multinomial coefficients.

But since we already did the case analysis and confirmed it, and the multinomial coefficients seem correct, maybe 16 is the right answer.

Wait, but let me recall that in the past, similar AIME problems have answers larger than 1000, but since this problem asks for N mod 1000, even if N is bigger than 1000, the answer can be anything. However, in our case, 2016 is a four-digit number, mod 2016, 2016 - 1000*2 = 16. So 16.

Wait, but wait, hold on. Wait, let me check again the sum:

252 + 504 is 756; 756 + 1260 is 2016. Yes.

But let me check the multinomial coefficients again:

First triple (6,2,1):

9! / (6! 2! 1!) = 362880 / (720 * 2) = 362880 / 1440 = 252. Correct.

Second (5,3,1): 362880 / (120 * 6) = 362880 / 720 = 504. Correct.

Third (4,3,2): 362880 / (24 *6 *2) = 362880 / 288 = 1260. Correct.

Yes, so total is 2016.

Wait, but now I just thought: Wait, is there an error in interpreting C > V > S? Because the problem says \"the number of players who chose chocolate was greater than the number who chose vanilla, which was greater than strawberry\". So C > V > S. So we have considered that. So if C, V, S are the counts for chocolate, vanilla, strawberry, then yes, C > V > S.

But in the problem statement, does it specify that the counts correspond to the flavors in that order? That is, is chocolate the largest, then vanilla, then strawberry? Yes, exactly. So our approach is correct.

Therefore, the answer should be 16.

Wait, but just to be 100\% sure, maybe I made a mistake in missing the triples?

Wait, let's list all possible ordered triples (C, V, S) where C, V, S >=1 and C + V + S =9, and C > V > S.

We already found:

(6,2,1), (5,3,1), (4,3,2). Is there another?

Wait, let's suppose:

Suppose (5,4,0), but S must be >=1. Not okay.

(4,2,3): but S must be the smallest, so if order matters C > V > S, so (4,3,2) is the only.

Wait, another thought:

Wait, (7,1,1). But C > V > S requires all three distinct, so no.

Wait, (3,2,4). Again, ordering would be (4,2,2)? No.

Wait, if we list all possible partitions with distinct parts:

Wait, partitions of 9 into 3 distinct parts >=1:

The largest part is 6: 6,2,1

Largest part 5: 5,3,1

Largest part 4: 4,3,2.

That's it. So three partitions.

Therefore, no more triples.

Thus, the answer is indeed 16.

Wait, but wait. I found another resource online. Wait, no, I shouldn't do that. But according to my calculations, 16. Alternatively, maybe I made a mistake in the multinomial coefficients.

Wait, wait another check:

Wait, (6,2,1): The number of assignments is 9! / (6! 2! 1!) = (9*8*7*6!) / (6! 2*1) ) = (9*8*7)/2 = 504 / 2 = 252. Correct.

(5,3,1): (9! / (5! 3! 1!)) = (362880) / (120 * 6) = 362880 / 720 = 504. Correct.

(4,3,2): (9! / (4! 3! 2!)) = (362880) / (24*6*2) = 362880 / 288 = 1260. Correct.

Sum: 2016. Yes.

Thus, my conclusion is that the answer is 16.

Wait, but hold on! Wait, hold on. Wait, hold on. Wait, wait a second. Wait, the problem says \"the number of different assignments of flavors to players\". Wait, but in our problem, we have labeled players, right? So each assignment is unique. So multinomial coefficients are correct.

Alternatively, if the players were unlabeled, but no, in AIME problems, people are considered distinguishable. So multinomial coefficients are correct.

Wait, but let me check a smaller case. Suppose there are 3 players, flavors chocolate, vanilla, strawberry. Each has flavor, with C > V > S >=1. So C=2, V=1, S=0. But S must be >=1. Wait, impossible. Wait, so another case.

Wait, maybe the calculation is correct.

Alternatively, if in the original problem, the total was 6 players, with C > V > S >=1. Then triples would be (3,2,1). Assignments: 6! / (3! 2! 1!) = 720 / 12 = 60. Or another triple (4,1,1), but not descending. So only (3,2,1). So 60 assignments. Makes sense.

Therefore, confident that our approach is correct.

Thus, the answer is 16.

Final Answer
$$\boxed{16}$$
</think>\\

To solve the problem, we are given a team of 9 players, each of whom chooses one scoop of ice cream of either chocolate, vanilla, or strawberry. The goal is to count the number of valid assignments, $ N $, such that:

1. Each flavor is chosen by at least one person.\\
2. The number of people who chose chocolate is greater than the number who chose vanilla, which is greater than the number who chose strawberry.

We are then to compute the remainder when $ N $ is divided by 1000.

---

Step 1: Understand the Problem

We are to find the number of ways to assign 9 players to 3 flavors (chocolate, vanilla, and strawberry), such that:

\- Each flavor is chosen by at least one player.\\
\- The count of players choosing chocolate is greater than the count choosing vanilla, which is greater than the count choosing strawberry.

Let’s denote the number of players who chose chocolate, vanilla, and strawberry as $ C, V, S $ respectively. Then the constraints become:

$ C + V + S = 9 $ \\
$ C > V > S \geq 1 $

---

Step 2: Find All Valid Triples $ (C, V, S) $

We seek all integer triples $ (C, V, S) $ that satisfy the constraints above.

Let’s iterate through possible values of $ S $ (the smallest) and find corresponding $ V $ and $ C $:

\- If $ S = 1 $:\\
  \- $ V \geq 2 $, $ C \geq V + 1 $, and $ C + V = 8 $\\
    \- Try $ V = 2 $ → $ C = 6 $ → $ (6, 2, 1) $  \\
    \- Try $ V = 3 $ → $ C = 5 $ → $ (5, 3, 1) $  \\
    \- $ V = 4 $ → $ C = 4 $ → $ C = V $ invalid

\- If $ S = 2 $:\\
  \- $ V \geq 3 $, $ C \geq V + 1 $, and $ C + V = 7 $\\
    \- Try $ V = 3 $ → $ C = 4 $ → $ (4, 3, 2) $  \\
    \- $ V = 4 $ → $ C = 3 $ → $ C < V $ invalid\\

\- If $ S = 3 $:\\
  \- $ V \geq 4 $, $ C \geq 5 $, and $ C + V = 6 $\\
    \- $ C + V \geq 9 $ which contradicts $ C + V = 6 $ → no solutions\\

So the only valid triples are:

$ (6, 2, 1) $\\
$ (5, 3, 1) $\\
$ (4, 3, 2) $\\

---

Step 3: Compute the Number of Valid Assignments

Each triple $ (C, V, S) $ corresponds to a multiset of assignments of the 9 players to the 3 flavors, with exactly $ C $ players choosing chocolate, $ V $ choosing vanilla, and $ S $ choosing strawberry.

The number of such assignments is given by the multinomial coefficient:

$$
\frac{9!}{C! V! S!}
$$

Let’s compute this for each triple:

1. $ (C, V, S) = (6, 2, 1) $:
   $$
   \frac{9!}{6! \cdot 2! \cdot 1!} = \frac{362880}{720 \cdot 2} = \frac{362880}{1440} = 252
   $$

2. $ (C, V, S) = (5, 3, 1) $:
   $$
   \frac{9!}{5! \cdot 3! \cdot 1!} = \frac{362880}{120 \cdot 6} = \frac{362880}{720} = 504
   $$

3. $ (C, V, S) = (4, 3, 2) $:
   $$
   \frac{9!}{4! \cdot 3! \cdot 2!} = \frac{362880}{24 \cdot 6 \cdot 2} = \frac{362880}{288} = 1260
   $$

Now, sum the number of assignments:

$$
N = 252 + 504 + 1260 = 2016
$$

---

Step 4: Compute the Remainder Mod 1000

$$
N \mod 1000 = 2016 \mod 1000 = \boxed{16}
$$

---

Final Answer

$$
\boxed{16}
$$


\end{tcolorbox}

\end{document}